\documentclass[letterpaper,10pt,conference]{ieeeconf}

\IEEEoverridecommandlockouts
\usepackage{amsmath,amssymb,amsfonts}
\usepackage{cite}
\usepackage{graphicx}
\usepackage{microtype}
\usepackage{url}
\usepackage{xcolor}
\usepackage{algorithm}
\usepackage{algpseudocode}
\usepackage{subcaption}
\usepackage{multirow}
\usepackage{booktabs}
\graphicspath{{figures/}}

\newtheorem{definition}{Definition}
\newtheorem{assumption}{Assumption}
\newtheorem{proposition}{Proposition}
\newtheorem{theorem}{Theorem}

\newcommand{\R}{\mathbb{R}}
\newcommand{\Sone}{\mathbb{S}^{1}}
\newcommand{\Cspace}{\mathcal{C}}
\newcommand{\Cfree}{\mathcal{C}_{\mathrm{free}}}
\newcommand{\CRC}{\mathcal{C}_{\mathrm{RC}}}

\newcommand{\Gmap}{\mathcal{G}}
\newcommand{\Roadmap}{\mathcal{M}_{\mathrm{RC}}}
\newcommand{\Render}{\mathcal{R}_{\Gmap}}

\newcommand{\Obstacles}{\mathcal{O}}
\newcommand{\Cover}{\mathcal{S}}

\newcommand{\RCValid}{\operatorname{RCValid}_{\Gmap}}
\newcommand{\BackCull}{\operatorname{BFCull}_{\Gmap}}
\newcommand{\method}{GS-PRM}

\title{\LARGE \bf
GS-PRM: Gaussian Visual Roadmap for Image-Goal Navigation}

\author{Ashraf Siddique, Kyung Min Han, Young J. Kim$^{*}$%
\thanks{$^{*}$Corresponding author: Young J. Kim.}}

\begin{document}
\maketitle
\thispagestyle{empty}
\pagestyle{empty}

%%%%%%%%%%%%%%%%%%%%%%%%%%%%%%%%%%%%%%%%%%%%%%%%%%%%%%%%%%%%%%%%%%%%%%%%%%%%%%%%
\begin{abstract}
Image-goal navigation requires a global planner that connects visually specified endpoints while providing observations that a local policy can follow. Three-dimensional Gaussian Splatting (3DGS) provides map-registered RGB, opacity, and depth renderings, but does not explicitly encode navigable topology. We introduce \method, a Gaussian visual roadmap whose milestones, local connections, and visual references are constructed directly through 3DGS rendering without deriving an occupancy grid or signed distance field. We define render-based collision-free (RC-free) configurations and paths. Panoramic depth supplies conservative clearance for milestone validation and adaptive coverage, while edge-aligned orthographic rendering tests the robot's swept volume. 
We establish conditional soundness and sufficient conditions for deterministic query completeness and probabilistic construction completeness of GS-PRM.
To validate the performance of GS-PRM, we conduct experiments in both simulated and real-world environments, demonstrating long-horizon navigation using goal-conditioned policies guided by GS-PRM. Compared with a state-of-the-art 3DGS-based navigation system, our GS-PRM-based navigation system achieves up to a twice higher average success rate and reduces travel time by up to 57.2\% when both systems complete the navigation task.

\end{abstract}

%%%%%%%%%%%%%%%%%%%%%%%%%%%%%%%%%%%%%%%%%%%%%%%%%%%%%%%%%%%%%%%%%%%%%%%%%%%%%%%%
\section{Introduction}

Image-goal navigation is a visual navigation problem that requires an agent to navigate to the location represented by a given goal image \cite{9981090}.  The image conveys appearance, but neither the goal pose nor a traversable route, and the destination is usually outside the current camera view. A goal-conditioned visual policy can reliably act only toward a nearby, recognizable observation within its training and control horizon. Long-horizon navigation, therefore, requires a persistent global topology that connects the localized query images and decomposes the route into reachable visual subgoals. It must also provide map-registered observations for repeated pose correction, since odometric drift can otherwise cause the physical robot to deviate from the planned route.
This task requires two forms of persistent scene information: a global planner needs the connectivity of traversable space, whereas the online robot needs map-registered visual observations for localization and feedback. 

{Probabilistic roadmaps (PRMs) construct a reusable graph by sampling collision-free milestones and connecting them through geometrically validated local paths~\cite{kavraki1996probabilistic,lavalle2006planning}. However, conventional PRMs assume sufficiently accurate workspace geometry for collision checking, which can require many expensive collision tests. For physical robots, reconstruction and localization errors can further invalidate these geometric tests. Moreover, conventional PRMs lack the visual observations needed for image-based localization and feedback.}

Three-dimensional Gaussian Splatting (3DGS) is attractive in this setting because it represents metric, map-registered appearance and supports fast rendering of RGB, opacity, and depth from arbitrary viewpoints~\cite{kwon2023renderable}. These renderings can provide visual references for localization and geometric evidence for planning through a common scene representation.
However, since raw Gaussian primitives are optimized for view synthesis and do not directly correspond to physical obstacles, a render-native planner must explicitly define what its renderer certifies. %However, raw Gaussian primitives are optimized for view synthesis and do not individually correspond to physical obstacles. A useful render-native planner must therefore define precisely what its renderer certifies instead of informally equating Gaussian primitives, rendered depth, and physical collision geometry.

Existing Gaussian-map planners take several approaches to this problem. SplatNav performs collision checking against Gaussian confidence ellipsoids and constructs safe volumes~\cite{chen2025splat}. GaussNav and related systems derive voxel, bird's-eye-view, or other planning maps from a Gaussian scene~\cite{lei2025gaussnav,chopra2025splatblox}. Other work directly renders visual subgoals for a learned navigation policy~\cite{honda2025gsplatvnm}. These methods demonstrate that 3DGS is useful for navigation. Still, they do not provide a persistent coverage-induced roadmap in which the same rendering interface defines milestone validity, local connectivity, image queries, and visual route references.

We address these limitations with \emph{\method}, a coverage-driven Gaussian visual roadmap. PRM provides a persistent multi-query topology, while 3DGS provides map-registered observations that are absent from a conventional roadmap. 
Using a single rendering interface, \method{} evaluates milestones and local connections, attaches image-defined endpoints, and generates visual references along the selected route.
We introduce the concept of {\em render-based collision-free} (RC-free), a conservative notion of free configuration space under some rendering assumptions,
providing a relationship between rendered evidence and planning. Clearance-adaptive coverage then reduces redundant
milestones in open areas while retaining finer resolution near constrained
regions.
%\blue{A challenge is that guaranteeing geometric collision-freeness based on standard 3DGS rendering is nontrivial because 3DGS does not provide an explicit geometric representation for collision checking. To address this problem, GS-PRM constructs a render-based collision-free (RC-free) graph directly in the robot configuration space (C-space), providing a relaxed notion of collision-freeness that is practical for real-world navigation.}
%\blue{Moreover, GS-PRM employs clearance-adaptive coverage that adjust the sampling density according to spatial clearance.}

% Candidate positions are sampled uniformly over the full workspace; samples that are already represented or fail the RC-free test are rejected. Thus every positive-measure uncovered RC-free region retains nonzero sampling probability. For robust RC-free paths and a locally complete connection rule, the probability of failing to connect a feasible query approaches zero as the sampling budget grows. Separately, if certified regions cover an RC-free component and their graph is connected, every attached query in that component has a roadmap path.

The contributions of this work are:
\begin{itemize}
    \item We construct a Gaussian visual roadmap from the same map-registered 3DGS renderings used for planning, localization, and image-goal guidance, without deriving a separate occupancy or distance-field planning map.
    \item We define RC-free milestones and local connections using conservative, thresholded depth, and state the assumptions under which their accepted configurations and paths lie within physical free space.
    \item We construct a connected roadmap from certified clearance-adaptive regions and establish sufficient conditions for probabilistic construction completeness and deterministic query completeness of a completed roadmap.
    \item For long-horizon visual navigation, we integrate image-based endpoint localization, validated roadmap attachment, and rendered visual subgoals for a goal-conditioned local policy.
    % \item We evaluate collision-oracle soundness, roadmap efficiency, and end-to-end navigation in simulation and on a physical robot.
    % \item{ We conducted extensive experiments in both simulated and real-world environments, demonstrating that GS-PRM-guided local planners substantially outperform their standalone counterparts. In particular, goal-conditioned planners combined with GS-PRM achieve 50\% success rates, whereas the same planners without GS-PRM achieve 0\% success rates. Compared with another GS-based planner, SplatNav, our GS-PRM-guided local planner also achieves a 50\% higher success rate and a 57.2\% reduction in travel time.}

    \item{We conducted a diverse set of experiments in both simulated and real-world 
    environments, demonstrating the effectiveness of GS-PRM for long-horizon 
    navigation. In particular, DevGRU combined with GS-PRM achieves a 50\% 
    success rate, whereas standalone DevGRU achieves a 0\% success rate. 
    Compared with another 3DGS-based planner, SplatNav, our GS-PRM combined with straight-line planner (SLP) also 
    achieves up to twice higher success rate and a 57.2\% reduction 
    in travel time.}

\end{itemize}

%%%%%%%%%%%%%%%%%%%%%%%%%%%%%%%%%%%%%%%%%%%%%%%%%%%%%%%%%%%%%%%%%%%%%%%%%%%%%%%%
\section{Related Work}

\subsection{Probabilistic Roadmaps for Navigation}

The classical PRM separates an offline learning phase, in which sampled milestones and locally feasible connections approximate free-space connectivity, from an online query phase~\cite{kavraki1996probabilistic}. Uniform, obstacle-based, bridge, Gaussian, visibility, and deterministic sampling strategies trade construction cost against narrow-passage discovery and dispersion; an overview is provided by LaValle~\cite{lavalle2006planning}. PRM variants have subsequently been adapted to navigation. T-PRM predicts the temporal availability of roadmap vertices in the presence of moving obstacles~\cite{HupChl22}. PRM-RL accepts edges by rolling out a learned short-range policy rather than relying solely on a straight-line local planner~\cite{FauTap18}. %Disk-Graph PRM associates milestones with free-space disks and exploits local clearance to guide exploration~\cite{NoeCha22}.

These methods exploit metric relationships among configurations but normally assume a geometric or state representation on which collision checking is defined. \method{} instead defines roadmap validity and coverage through 3DGS rendering and associates the resulting roadmap with visual observations for online navigation.

\subsection{3D Gaussian Splatting for Robot Navigation}

3DGS combines an explicit set of spatial primitives with fast novel-view
synthesis~\cite{kerbl20233d}. Gaussian Splatting SLAM demonstrates that a Gaussian scene can jointly support online mapping, camera tracking, and novel-view rendering, while GSplatLoc estimates camera pose by aligning observed and rendered depth~\cite{matsuki2024gaussian,zeller2025gsplatloc}. These works
establish 3DGS as a map-registered visual representation, but do not supply the persistent traversability graph required for multi-query planning.

Gaussian-map planners differ in how they obtain collision geometry. GaussNav constructs a semantic Gaussian map for image-goal matching and converts it to a grid for path planning~\cite{lei2025gaussnav}; SplatBlox similarly derives a planning-oriented spatial representation~\cite{chopra2025splatblox}. SplatNav instead checks collisions against Gaussian confidence ellipsoids and constructs safe volumes~\cite{chen2025splat}.
%, whereas FOCI optimizes orientation-aware trajectories using overlap integrals between robot and scene Gaussians~\cite{andreu2025foci}. 
These methods either derive auxiliary planning geometry or reason directly about Gaussian primitives.

Rendering has also been used as the interface to a visual policy. GSplatVNM synthesizes intermediate point-of-view images that bridge gaps in a sparse image database~\cite{honda2025gsplatvnm}, and 3DGSNav renders informative views from a persistent Gaussian memory for long-horizon object navigation~\cite{zheng2026gsnav}. In contrast, \method{} builds a reusable coverage roadmap whose milestone test, connection test, image-query interface, and visual route references are all defined through rendering.

Rendered depth is not automatically a conservative collision oracle. It may be affected by missing surfaces, low accumulated opacity, depth blending, thin objects, finite raster resolution, and view dependence. We therefore make the relationship between rendered evidence and physical collision freedom an explicit assumption and evaluate it separately from roadmap connectivity.

\subsection{Goal-Conditioned Visual Navigation}

Image-goal policies infer short-horizon motion or waypoints from the current observation and a goal image. GNM learns this interface from heterogeneous robot datasets, and ViNT scales it into a reusable navigation model that generalizes across environments and embodiments~\cite{shah2023gnm, shah2023vint}. NoMaD further models multimodal action sequences using a goal-masked diffusion policy and supports both goal-reaching and exploration~\cite{sridhar2024nomad}. Related approaches combine learned local control with topological memory or high-level goal proposal for longer-horizon navigation~\cite{hatch2025ghil,Yang-RSS-23}.

These policies remain local and require a reachable, visually informative goal. \method{} supplies a render-validated global graph in which pose-indexed observations serve as subgoals and localization references, extending the policy horizon without requiring a densely recorded image graph.

\begin{figure*}[t]
    \centering
    \IfFileExists{figures/pipeline_simple1_db.pdf}{%
        \includegraphics[width=1.0\linewidth]{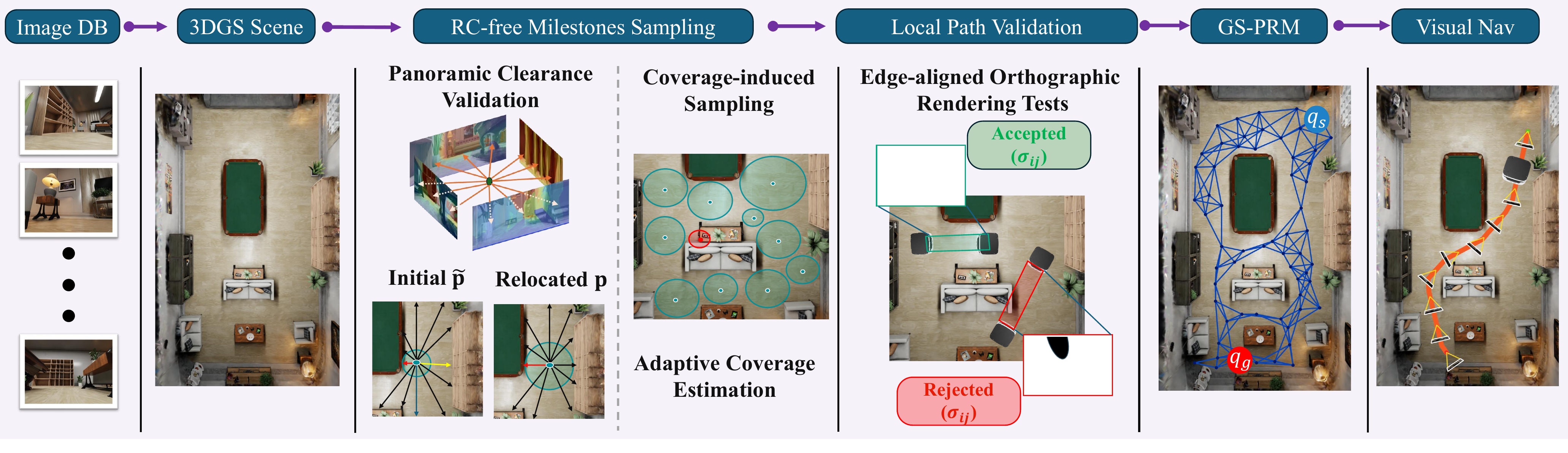}%
    }{%
        \fbox{\parbox[c][31mm][c]{0.94\linewidth}{\centering
        Placeholder for the \method{} pipeline: 3DGS rendering $\rightarrow$
        RC-free coverage and connected roadmap $\rightarrow$ image-query
        attachment $\rightarrow$ graph search and visual subgoals.}}%
    }
    \caption{\method{} constructs a persistent coverage-induced roadmap
    directly from 3DGS rendering. The same scene representation supports
    render-based milestone and edge tests, image-query localization, and visual
    subgoal generation for online navigation.}
    \label{fig:pipeline}
\end{figure*}

%%%%%%%%%%%%%%%%%%%%%%%%%%%%%%%%%%%%%%%%%%%%%%%%%%%%%%%%%%%%%%%%%%%%%%%%%%%%%%%%
\section{Problem Formulation and Render-Based Free Space}
\label{sec:problem}

\subsection{Image-Goal Navigation Query}

% We consider a mobile robot moving on a known horizontal support surface. Its
% planar configuration is
% \begin{equation}
%     q=(\mathbf p,\psi)\in\Cspace=\Omega\times\Sone,
%     \label{eq:configuration}
% \end{equation}
% where $\mathbf p\in\Omega\subset\R^2$ is the robot reference position and
% $\psi$ is its heading. The physical robot body at $q$ is $\Body(q)\subset
% \R^3$, and the physical obstacle set is $\Obstacles$. The true free
% configuration space, used only for analysis, is
% \begin{equation}
%     \Cfree=\{q\in\Cspace\mid \Body(q)\cap\Obstacles=\emptyset\}.
%     \label{eq:true_free}
% \end{equation}
% We conservatively enclose the chassis by a circular footprint, which makes collision validity independent of heading and permits clearance to be compared with a single radius. Heading remains part of the state because it determines rendered observations and local-policy commands. A navigation query consists of start and goal images $(I_s,I_g)$. The objective is to localize their configurations $(q_s,q_g)$ and return a physically feasible path between them, if both lie in the same navigable component.

We consider a mobile robot operating on a known horizontal support surface. Its planar configuration is
\begin{equation}
q=(\mathbf p,\psi)\in\Cspace=\Omega\times\Sone,
\label{eq:configuration}
\end{equation}
where $\mathbf p\in\Omega\subset\R^2$ denotes the robot reference position and $\psi$ its heading. 
%Let $\Body(q)\subset\R^3$ denote the physical volume occupied by the robot at configuration $q$, and 
Let $\Obstacles$ denote the set of physical obstacles. The corresponding free configuration space is denoted as $\Cfree$.
%\begin{equation}
%\Cfree={q\in\Cspace\mid %\Body(q)\cap\Obstacles=\emptyset}.
%\label{eq}
%\end{equation}
By conservatively bounding the robot chassis with a circular footprint, collision validity becomes independent of heading.
Nevertheless, $\psi$ remains a necessary component of the configuration space because the robot's heading determines both the rendered observations for the local navigation policy and the local policy command. An image-goal navigation query consists of a start and a goal image, $(I_s,I_g)$. The primary objective is to localize their respective configurations $(q_s, q_g)$ and estimate a physically feasible, navigable path.

\subsection{RC-free Configuration and Path}

Let $\Gmap$ be a preconstructed 3DGS scene. Given a calibrated camera pose $T_c$ and camera intrinsic parameters $\mathrm{K}$, the rendering function $\Render$ produces RGB $I$, depth $D$, and accumulated opacity $A$:
\begin{equation}
    (I,D,A)=\Render(T_c,\mathrm{K}).
    \label{eq:renderer}
\end{equation}
For each rendered ray $u$, $\widehat d_{\Gmap}(u)$ denotes the first depth at which the accumulated opacity reaches $\tau_\alpha\in(0,1)$. If no threshold crossing occurs, the ray is treated as free only over a sufficiently render-supported interval; otherwise, it is marked as unknown. For a known ray with no threshold crossing over the tested interval, we set $\widehat d_{\Gmap}(u)=+\infty$ for the corresponding free-space predicate. In contrast, a ray whose tested interval lacks sufficient support is labeled unknown rather than free.
For a configuration $q$, a panoramic render defines a set of rays $\mathcal U_q$ that conservatively covers the safety-inflated robot height and footprint in all horizontal directions. Let $r_{\mathrm{safe}}$ include the robot radius and a fixed safety clearance margin.

\begin{definition}[RC-free configuration]
A configuration $q$ is \emph{render-based collision-free} (RC-free),
denoted $q\in\CRC$, if and only if its entire safety-inflated robot
body has sufficient reconstruction support\footnote{This means that the 3DGS scene is adequately reconstructed to provide reliable depth and opacity under Assumption \ref{assump:opaque}.} and every robot-relevant
panoramic ray satisfies
\begin{equation}
    \widehat d_{\Gmap}(u)\ge r_{\mathrm{safe}}
    \qquad \forall u\in\mathcal U_q,
    \label{eq:rc_free_configuration}
\end{equation}
where $\mathcal U_q$ is the set of panoramic rays covering the robot
body, and $r_{\mathrm{safe}}$ includes the robot radius and an
additional clearance margin.
\end{definition}

For a straight local path $\sigma_{ij}$ from $q_i$ to $q_j$,
let $L_{ij}$ denote its translational length. An edge-aligned
orthographic render provides a ray set $\mathcal U_{ij}$ covering
the complete safety-inflated cross-section of the swept volume.
The minimum thresholded depth over these rays is
$\widehat d_{ij}^{\min}
=\min_{u\in\mathcal U_{ij}}\widehat d_{\Gmap}(u)$.

\begin{definition}[RC-free local path]
A local path $\sigma_{ij}$ connecting $q_i, q_j\in \CRC$ is \emph{RC-free}, denoted
$\RCValid(\sigma_{ij})=1$, if and only if its entire safety-inflated
swept volume has sufficient reconstruction support and
\begin{equation}
    \widehat d_{ij}^{\min}\ge L_{ij}+\delta_e,
    \label{eq:rc_free_edge}
\end{equation}
where $\delta_e\ge0$ is a rendering and execution margin.
Thus, no ray in $\mathcal U_{ij}$ detects a thresholded surface before the
required clearance distance. %{Configurations and local paths with insufficient reconstruction support are labeled as unknown.}
\end{definition}
% Configurations and local paths with insufficient reconstruction support
% are rejected; unknown regions are never treated as free.

%Unlike a percentile rule, Eq.~\eqref{eq:rc_free_edge} rejects an edge when even one conservatively rasterized corridor ray observes an obstacle before the endpoint. This worst-case predicate is important for thin obstacles. In the discrete implementation, the rendered obstacle mask is dilated by a pixel-scale bound before the minimum depth is evaluated.

\subsection{Collision-free Conditions}

%Opacity of physical objects alone does not guarantee correct rendered depth. 
We state the assumptions required to relate RC-free $\CRC$ and RC-free path $\sigma$ to free configuration space $\Cfree$.

\begin{assumption}[Opaque-surface depth]
%The environment is static, and every physical obstacle surface is opaque. 
Whenever a tested ray intersects an obstacle, the 3DGS representation accumulates opacity of at least $\tau_\alpha$ before the true first-surface depth. 
%A sufficient theoretical certificate requires, for every tested ray that intersects an obstacle, $\widehat d_{\Gmap}(u)\le d_{\Obstacles}(u)$, where $d_{\Obstacles}$ is the true first-intersection depth. 

% In practice, this inequality can be checked against held-out depth or LiDAR and uncertified rays can be rejected.
\label{assump:opaque}
\end{assumption}

\begin{assumption}[Conservative ray coverage]
The panoramic and orthographic ray sets are designed to cover the safety-inflated robot body and its swept volume conservatively. Areas with insufficient reconstruction support are treated as unknown and are not considered free space. 
%For a reliable discrete test, the ray spacing and mask dilation should be selected so that every obstacle above a minimum specified size is detected by at least one ray. In addition, the entire tested volume must satisfy Eq.~\eqref{eq}.
\label{assump:rays}
\end{assumption}

% \begin{proposition}[Conditional soundness]
% \label{prop:soundness}
%  Under Assumptions~\ref{assump:opaque} and~\ref{assump:rays},
% \begin{equation}
% \begin{gathered}
%     \CRC\subseteq\Cfree,\\[-1mm]
%     \sigma_{ij}(0)=q_i, \sigma_{ij}(1)=q_j, 
%     \RCValid(\sigma_{ij})=1\ \Longrightarrow\
%     \sigma_{ij}([0,1])\subseteq\Cfree.
% \end{gathered}
%     \label{eq:soundness}
% \end{equation}
% \end{proposition}

\begin{proposition}[Conditional soundness]
\label{prop:soundness}
Under Assumptions~\ref{assump:opaque} and~\ref{assump:rays},
\begin{equation}
\begin{aligned}
    \CRC &\subseteq \Cfree \quad \textrm{and}\\[-0.5mm]
    \RCValid(\sigma_{ij})=1
    &\Longrightarrow
    \sigma_{ij}([0,1])\subseteq\Cfree.
\end{aligned}
\label{eq:soundness}
\end{equation}
\noindent
where $\sigma_{ij}:[0,1]\rightarrow\mathcal{C}$,
$\sigma_{ij}(0)=q_i$, and $\sigma_{ij}(1)=q_j$.
\end{proposition}

%\vspace{2mm}

% \begin{remark}
% Equation~\eqref{eq:soundness} is a conditional soundness statement, not an unconditional property of standard 3DGS. Reconstruction holes, depth overestimation, insufficient opacity, or missed subpixel obstacles can violate Assumptions~\ref{assump:opaque} and~\ref{assump:rays}. The false-free rate of the oracle must therefore be evaluated against geometric ground truth.
% \end{remark}

%%%%%%%%%%%%%%%%%%%%%%%%%%%%%%%%%%%%%%%%%%%%%%%%%%%%%%%%%%%%%%%%%%%%%%%%%%%%%
{Fig.~\ref{fig:pipeline} provides an overview of the proposed GS-PRM framework, including roadmap construction and its use for image-goal navigation, which are discussed in the following sections.}

\vspace{-4pt}
\section{Coverage-Induced Roadmap}
\label{sec:roadmap}

\subsection{GS-PRM}

Let $\widehat c_i$ be the minimum thresholded panoramic depth at $q_i$. 
For a circular robot, the translational coverage radius is
\begin{equation}
    \rho_i=\min\{\beta(\widehat c_i-r_{\mathrm{safe}}),\rho_{\max}\},
    \qquad 0<\beta\le1,
    \label{eq:coverage_radius}
\end{equation}
{where $\rho_{\max}$ is the maximum allowed coverage radius}, and the associated coverage region is
\begin{equation}
    \Cover_i=B(\mathbf p_i,\rho_i)\times\Sone,
    \label{eq:coverage_region}
\end{equation}
where
$B(\mathbf p,r)=\{\mathbf x\in\Omega:\|\mathbf x-\mathbf p\|_2\le r\}$.
%denotes the closed planar disk of radius $r$ centered at $\mathbf p$. 
In open areas, larger coverage regions help reduce redundant milestones. In narrow or constrained areas, smaller regions allow more samples to be placed.
Under {Assumption~\ref{assump:opaque}, $\widehat c_i$ gives a conservative estimate of the distance from the robot center to the nearest physical obstacle. Since this distance can decrease by at most the amount the robot moves, Eq.~\eqref{eq:coverage_radius} ensures that every position inside $\Cover_i$ still maintains the required safety clearance. Therefore, $\Cover_i$ is used not only to avoid redundant sampling but also to represent a locally certified free region.

\begin{definition}[GS-PRM]
A \method{} roadmap is a graph
\begin{equation}
    \Roadmap=(V,E),
    \label{eq:roadmap}
\end{equation}
where each milestone
\begin{equation}
    v_i=(q_i,\Cover_i,\Phi_i)\in V
    \label{eq:milestone}
\end{equation}
contains an RC-free configuration $q_i \in \CRC$, its certified local coverage region $\Cover_i$, and map-registered visual attributes $\Phi_i=(I_i,D_i,A_i,F_i)$, where $F_i$ denotes optional image feature descriptors. Each edge $(v_i,v_j)\in E$ represents a local path $\sigma_{ij}$ satisfying $\RCValid(\sigma_{ij})=1$. Images and descriptors may be stored offline or rendered on demand during online navigation.
\end{definition}

\subsection{Coverage-Guided Sampling}
Since the translational footprint of $\Cover_i$ is simply $B(\mathbf p_i,\rho_i)$, we therefore define the currently uncovered sampling domain:
\begin{equation}
    U_n=\Omega\setminus\bigcup_{i=1}^{n}B(\mathbf p_i,\rho_i).
    \label{eq:uncovered_domain}
\end{equation}
At every trial $\ell$, the construction draws a position independently and uniformly over the entire workspace,
\begin{equation}
    \widetilde{\mathbf p}^{(\ell)}\sim\operatorname{Unif}(\Omega).
    \label{eq:uniform_sampling}
\end{equation}
The proposal is rejected without insertion if
$\widetilde{\mathbf p}^{(\ell)}\notin U_n$.
%, because it is already covered.

%For an uncovered proposal $q$, we first apply a low-cost \emph{back-face interior prefilter} using a rendered panoramic RGB and normal.
For an uncovered proposal $q$, we first apply a low-cost \emph{back-face interior prefilter} using a panoramic RGB and normal render \cite{akenine2018real}. A viewpoint inside an opaque surface typically observes a large fraction of nearly black pixels, due to one-sided appearance or surface normals facing away from the viewpoint.  Let $\mathcal V_q$ be the supported rendered pixels, $\mathcal{L}(I_q(u))\in[0,1]$ their luminance, $\mathbf n_q(u)$ the rendered outward normal, and $\mathbf v_q(u)$ the unit vector from the rendered surface point toward the camera. We compute
 \begin{equation}
 \begin{aligned}
 \gamma_{\mathrm{dark}}(q)
 &=\frac{1}{|\mathcal V_q|}\sum_{u\in\mathcal V_q}
 \mathbf 1[\mathcal{L}(I_q(u))<\tau_I],\\
 \gamma_{\mathrm{back}}(q)
 &=\frac{1}{|\mathcal V_q|}\sum_{u\in\mathcal V_q}
 \mathbf 1[\mathbf n_q(u)^\top\mathbf v_q(u)<-\tau_n],
 \end{aligned}
 \label{eq:backface_scores}
 \end{equation}
 and reject the proposal when
 \begin{equation}
 \begin{aligned}
 \BackCull(q)=1\iff{}&
 \gamma_{\mathrm{dark}}(q)>\tau_{\mathrm{dark}}\\
 &{}\lor\ \gamma_{\mathrm{back}}(q)>\tau_{\mathrm{back}}.
 \end{aligned}
 \label{eq:backface_cull}
 \end{equation}
 Either cue may be disabled when reliable colors or normals are unavailable. Proposals with insufficient supported pixels are rejected as unknown. %This test supplies negative evidence only: passing it does not establish that a configuration is free or RC-free.
If the surfaces reconstructed from $\mathcal{G}$ are opaque, as in Assumption \ref{assump:opaque}, and orientable, the back-face prefilter effectively determines RC-free configurations.
An uncovered proposal that passes this prefilter is tested by Eq.~\eqref{eq:rc_free_configuration}; a non-RC-free proposal is also rejected. 
Therefore, only uncovered RC-free proposals proceed to relocation and graph connection. Consequently, adaptive coverage changes which samples are retained, but it does not remove uniform proposal support {over $\Omega$.} %from any part of $\Omega$.

Before insertion, an uncovered RC-free proposal may be relocated as follows:
\begin{enumerate}
\item At iteration $t$, render the supported thresholded depths
$d_k^{(t)}$ along directions $\theta_k=2\pi k/M$, with
$\mathbf e_k=[\cos\theta_k,\sin\theta_k]^\top$, and assign greater weight to nearby surfaces with distance-decay parameter $\lambda_d>0$:
\begin{equation}
\begin{aligned}
\delta_k^{(t)}&=\max\{d_k^{(t)}-r_{\mathrm{safe}},0\},\\
w_k^{(t)}&=\exp\!\left(-\delta_k^{(t)}/\lambda_d\right),
\end{aligned}
\label{eq:relocation_weight}
\end{equation}
\item Compute their normalized repulsion and update the candidate position:
\begin{equation}
\mathbf F^{(t)}=-\frac{\sum_{k=1}^{M}w_k^{(t)}\mathbf e_k}
{\sum_{k=1}^{M}w_k^{(t)}},\qquad
\mathbf p^{(t+1)}=\mathbf p^{(t)}+\eta\mathbf F^{(t)}.
\label{eq:relocation_update}
\end{equation}
\item Rerender after every update and retain only iterates in $U_n\cap\CRC$. To preserve the sampling argument below, a relocated iterate is admissible only if its coverage disk contains the original proposal $\widetilde{\mathbf p}^{(\ell)}$; otherwise, the unrelocated proposal is retained. Stop when $\|\mathbf F^{(t)}\|$ is small or after a fixed budget, and select the admissible iterate with the largest minimum directional clearance. \end{enumerate} Relocation can therefore enlarge the subsequent coverage region without changing the RC-free acceptance test or discarding the point supplied by the uniform proposal process.
Note that this procedure is not a maximal-clearance roadmap~\cite{lavalle2006planning} utilizing the medial-axis structure to maximize obstacle clearance. In \method{}, candidates are still generated from the full workspace, but only uncovered candidates are retained, and relocation is only a bounded local heuristic for reducing redundant milestones. 
%It neither claims a clearance optimum nor determines graph connectivity; the final milestone, coverage region, and every connection are independently revalidated. Consequently, relocation affects finite-budget efficiency but is not used in the soundness or completeness arguments. 
% The theoretical coverage condition is exact set coverage, not merely an area ratio approaching one. For a connected component $\mathcal K\subseteq\CRC$, the required condition is
% \begin{equation}
%     \mathcal K\subseteq\bigcup_{i=1}^{|V|}\Cover_i.
%     \label{eq:exact_cover}
% \end{equation}
% A practical coverage ratio may be used as a finite-budget termination measure, but a ratio below one does not establish Eq.~\eqref{eq:exact_cover}; even a measure-one cover can omit a topologically important set.

\subsection{Graph Construction}

For each accepted candidate $v_i$, we propose neighboring milestones using coverage overlap and a maximum connection distance:
\begin{equation}
\mathcal N_i=
\left\{v_j\in V:\
\|\mathbf p_i-\mathbf p_j\|\le
\min(\rho_i+\rho_j,r_{\mathrm{conn}})\right\}.
\label{eq:neighbors}
\end{equation}
{where $r_\mathrm{conn}$ denotes the maximum connection distance, preventing connections between distant vertices.} Additional $k$-nearest candidates are tested to improve graph redundancy. Every proposed local path is evaluated by
Eq.~\eqref{eq:rc_free_edge}.

For each candidate pair $(v_i,v_j)$, the RC-valid edge test is implemented in four steps:
\begin{enumerate}
    \item Compute $L_{ij}=\|\mathbf p_j-\mathbf p_i\|_2$ and align an
    orthographic virtual camera at $\mathbf p_i$ with its viewing direction
    parallel to $\mathbf p_j-\mathbf p_i$.
    \item Set the orthographic image footprint to the vertical and lateral
    cross-section of the robot enlarged by the safety and execution margins.
    Render depth and accumulated opacity, and crop the robot-scale volume
    mask $\mathcal M_{ij}$; this mask supplies the ray set
    $\mathcal U_{ij}$ used by the RC-free local-path predicate.
    \item Convert the crop to thresholded depths using $\tau_\alpha$, mark
    unsupported pixels as unknown, and compute
    $\widehat d_{ij}^{\min}=\min_{u\in\mathcal M_{ij}}
    \widehat d_{\Gmap}(u)$.
    \item Accept the directed transition only when the volume is known and
    $\widehat d_{ij}^{\min}\ge L_{ij}+\delta_e$, as required by
    Eq.~\eqref{eq:rc_free_edge}. For an undirected edge, repeat the render from
    $v_j$ toward $v_i$ and require both directions to pass.
\end{enumerate}

The roadmap is initialized from one seed milestone in the navigable component of interest. A new milestone is committed only when it has at least one RC-valid edge to the existing roadmap; otherwise, it is deferred until the coverage frontier reaches it. {Following this procedure,} all RC-valid candidate edges are retained. 

\subsection{Completeness Guarantees}
One can show that under the standard PRM assumptions, the independent proposals $\widetilde{\mathbf p}^{(\ell)}\sim\operatorname{Unif}(\Omega)$ ensure that every measurable uncovered region $A\subseteq U_n$ with positive area has a nonzero probability of being sampled. For a robust RC-free component $\mathcal K\subseteq\CRC$, if $\mathcal K\subseteq\bigcup_i\Cover_i$ and neighboring milestones are connected through paths satisfying $\RCValid(\sigma_{ij})=1$, then any attached query configurations $q_s,q_g\in\mathcal K$ can be connected by an RC-free roadmap path. We only state the resulting theorems here.
%Its physical collision-freeness is conditional on Assumptions~\ref{assump:opaque} and~\ref{assump:rays}.}
%We distinguish completeness of a finished roadmap from the convergence of its random construction. 
%Theorem~\ref{thm:completeness} does not itself show that random construction
%attains exact coverage. 
%The next result connects the uniform proposal process
%in Eq.~\eqref{eq:uniform_sampling} to the usual probabilistic-completeness
%argument.

\begin{theorem}[Deterministic query completeness]
\label{thm:completeness}
For every $q_s,q_g\in\mathcal K$, where $\mathcal K$ is a connected component of $\CRC$, \method{} returns an RC-free path
from $q_s$ to $q_g$. Under Assumptions~\ref{assump:opaque} and
\ref{assump:rays}, the returned path is also collision-free.
\end{theorem}
\begin{theorem}[Probabilistic completeness of construction]
\label{thm:prob_complete}
\begin{equation}
\lim_{N_{\max}\rightarrow\infty}
\Pr\!\left[\text{\method{} fails to connect }q_s\text{ and }q_g\right]=0.\nonumber
%\label{eq:prob_complete}
\end{equation}
where \(N_{\max}\) denotes the maximum number of independent trials used during roadmap construction.
\end{theorem}

\section{Image-Defined Queries and Online Navigation}
\label{sec:navigation}

\subsection{Start and Goal Localization}

The 3DGS training images and calibrated poses form a map-registered retrieval database. For a query image, nearby views are retrieved and matched using SuperPoint~\cite{superpointdetone2018} and LightGlue~\cite{lightgluelindenberger2023}. Rendered depth is used to back-project reference pixels, and RANSAC-PnP~\cite{LepFua09-epnp} estimates the camera pose, which is then converted to the robot configuration using camera-to-base calibration.
The localized start and goal are temporary milestones. Each must satisfy the RC-free configuration predicate and be attached to the persistent graph via an RC-free local connection. %Multiple retrieved hypotheses are geometrically verified; a query is rejected when none passes the feature, reprojection, render-support, and roadmap-attachment tests.

\subsection{Graph Search and Route Validation}

Accepted edges retain length and optional execution-quality attributes. The standard A* search returns an optimal milestone sequence $\hat{\mathcal T}$, using geometric length and the normalized local coverage region as edge costs. Any shortcutting or spline smoothing is query-dependent and is accepted only after the complete modified trajectory is re-evaluated with the RC-free configuration and volume predicates. This prevents smoothing from leaving the volumes validated during roadmap construction.

The final smoothed route $\mathcal T$ is sampled according to the horizon of the navigation task. Pose-conditioned RGB and depth observations are rendered at the sampled poses, forming visual subgoals of a global path $\mathcal{T}_g \subset \mathcal T$ for the goal-conditioned policy. Thus, the visual information is operational rather than passive graph metadata: it defines image queries, supports online localization, and guides local execution.

\subsection{Visual Subgoal Execution and Online Robustness}

Let $\pi_\theta$ denote the goal-conditioned local navigation policy used for short-horizon execution~\cite{han2026devgru}. At time $t$, the policy receives the current observation $I_t$ and the next rendered roadmap observation $I_k^{\mathrm{ref}}$, optionally with depth $D_t, D_k^{\mathrm{ref}}$, and produces a short-horizon control command or waypoint, 
%\begin{equation}
%    a_t=\pi_\theta(I_t,I_k^{\mathrm{ref}}, D_t,D_k^{\mathrm{ref}}).
%    \label{eq:local_policy}
%\end{equation}
\begin{equation}
    a_t=\pi_\theta\!\left(
        I_t,I_k^{\mathrm{ref}};\,[D_t],[D_k^{\mathrm{ref}}]
    \right).
    \label{eq:local_policy}
\end{equation}

The roadmap, therefore, extends the operating horizon of a visual policy by replacing one distant image goal with a sequence of nearby, map-registered visual goals. Possible choices for $\pi_{\theta}$ include goal-conditioned navigation policies such as ViNT~\cite{shah2023vint} and DevGRU~\cite{han2026devgru}.

As an alternative to $\pi_\theta$, one can design a simpler, deterministic control policy, called the straight-line planner (SLP), where a PnP procedure is used to estimate the relative pose error between $I_t,I_k^{\mathrm{ref}}$ by matching their feature descriptors $F_t, F_k^{ref}$. The robot is then steered toward $I_k^{\mathrm{ref}}$ by minimizing the error.

% For long-horizon navigation with the off-the-shelf goal-conditioned local planners, the robot may correct its pose using the same PnP procedure at the end of each navigation segment to mitigate accumulated localization drift.

The same PnP procedure can also complement off-the-shelf goal-conditioned local planners, which may suffer from accumulated localization drift over long horizons. Therefore, applying pose correction at the end of each navigation segment can mitigate such drift and support long-horizon navigation.

\section{Experiments and Results}
\label{sec:experiments_and_results}

For the experiments, we employed a differential-drive mobile robot measuring $40\times40\times33 cm$ equipped with an Intel RealSense D435 RGB-D camera and a SICK TiM571 2D LiDAR sensor. %(see Fig.~\ref{fig:former_robot}). 
The 2D LiDAR was used solely to collect navigation data and to conduct post-analysis of each navigation run, and was not used for online navigation. The robot's onboard PC featured an AMD Ryzen 9 6900HX (1.6~GHz) CPU and an NVIDIA RTX 3070 GPU. All baseline models were integrated into ROS. We also use a discrete NVIDIA GeForce RTX 5090 GPU for computing GS-PRM and running Isaac Sim.

\subsection{Verifying Physical Soundness}
\label{subsect:simulation_experiments}
We first evaluated GS-PRM in an Isaac Sim environment.  First of all, experiments are designed to address the following two questions related to the physical soundness of GS-PRM:
\begin{enumerate}
\item[\textbf{Q1}] Is the RC-free condition a valid criterion for collision checking against ground-truth geometry?
\item[\textbf{Q2}] Is the trajectory planned by GS-PRM reliable for navigation? If so, how does its performance compare with the state-of-the-art approach?
\end{enumerate}

%The simulated robot was configured to match the mobile robot used in the real-world experiments. Images rendered from the simulation environment were used to reconstruct the 3DGS scene, which was then used for both roadmap construction and navigation. 
% The simulation experiments evaluate two aspects of the proposed method: the reliability of the 3DGS-based collision checker and the navigation performance of GS-PRM compared with SplatNav.

\subsubsection{Validation of RC-free Condition}
% We evaluated our collision checker using the known collision geometry available in Isaac Sim. The reconstructed 3DGS scene was used to classify uniformly sampled robot positions as RC-free or rejected. For ground-truth validation, the robot was placed at each evaluated position in the original simulation scene, and collisions were determined using the simulator geometry. Floor contacts below the specified collision-height threshold were ignored.

% A total of 654 samples were evaluated and shown in Fig.~\ref{fig:isaac_collision_validation}. All 131 samples accepted by the 3DGS checker were collision-free in Isaac Sim. Of the 523 rejected samples, 334 were confirmed collisions and 189 were collision-free. The latter mainly represent conservative rejections caused by the robot footprint and safety-clearance requirements rather than direct physical collision.

To answer Q1, we constructed a 3DGS representation of the Isaac Sim environment and uniformly sampled robot positions throughout the environment. We then evaluated each sampled position using our RC-free criterion. Among all sampled positions, 20.0\% were classified as RC-free, all of which were confirmed collision-free by the simulator's ground truth. Of the remaining 80\% rejected samples, 63.9\% corresponded to actual collisions, whereas 36.1\% were collision-free. These collision-free rejected samples primarily reflect conservative rejections due to the robot footprint and safety clearance constraints, as shown in Fig.~\ref{fig:isaac_collision_validation}. Thus, the RC-free criterion produced no false-negative classifications in this experiment, consistent with Proposition~\ref{prop:soundness}.
%, which states that, under Assumptions~\ref{assump:opaque} and~\ref{assump:rays}, $\CRC\subseteq\Cfree$.

\begin{figure}[t]
    \centering
    \includegraphics[width=\linewidth]
    {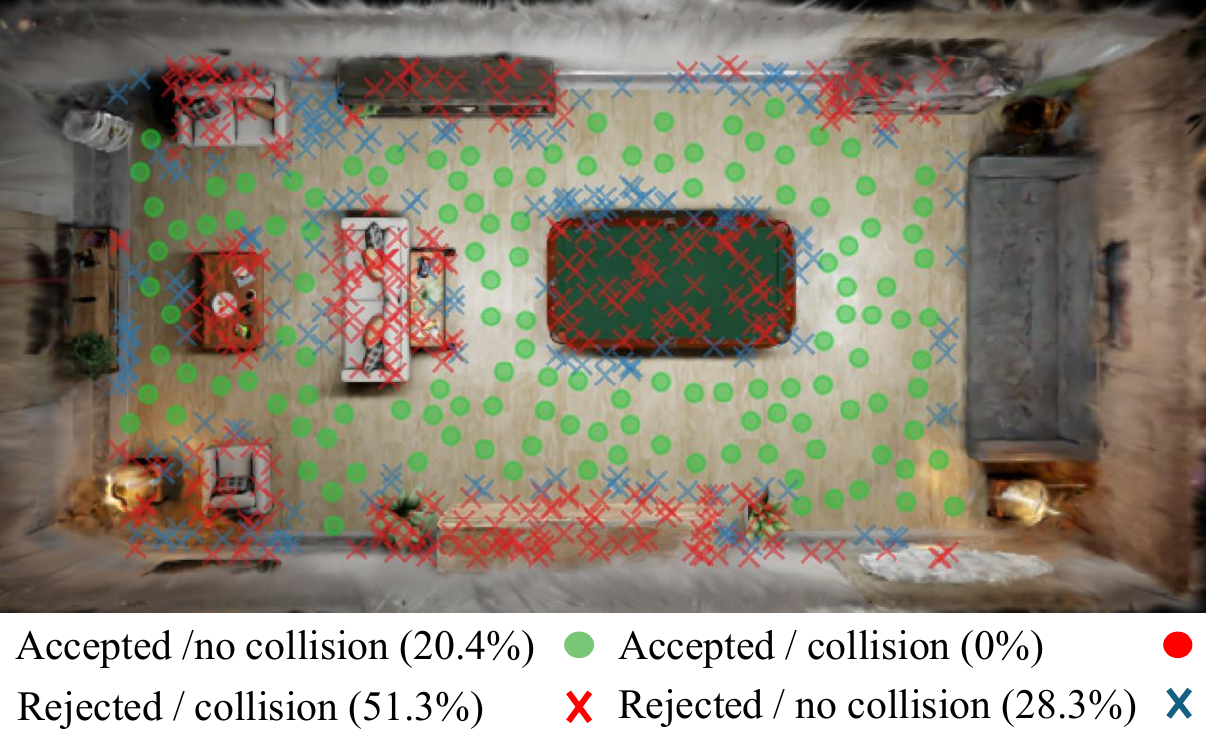}
    \caption{\textbf{Validation of the RC-free criterion in the simulated environment ($6.57 \times 14.1 \times 4.11\mathrm{m}$).}
    %Green circles denote RC-free samples that were confirmed to be collision-free,
    %red crosses denote rejected samples confirmed as
    %collisions, and blue crosses denote conservative rejections that were collision-free. No false-free classifications were observed in this experiment. 
    } %
    \label{fig:isaac_collision_validation}
\end{figure}
%\vspace{-8pt}

\subsubsection{Navigation Validation}
% To answer Q2, we compared the navigation performance of GS-PRM with that of SplatNav, our baseline. To this end, we constructed a GS-PRM in the aforementioned simulated environment and randomly selected pairs of start and goal vertices. For each pair, we computed a global trajectory using GS-PRM and then employed the straight-line local planner (SLP) to follow it. We then compared the resulting navigation performance with that of SplatNav. Table~\ref{tab:round1_navigation_comparison} reports the average navigation performance over 20 repeated trials for each experiment. %Fig.~\ref{fig:sim_navigation_compare} qualitatively illustrates a representative trial. 
% Our method outperformed SplatNav in terms of planned trajectory length, traveled path length, and navigation time by 5.16\%, 9.59\%, 17.47\%, respectively. One key factor contributing to this improvement is SplatNav's reliance on a coarse OGM for generating its initial trajectory.

{To answer Q2, we compared GS-PRM combined with SLP planner (GS-PRM+SLP) against  SplatNav using randomly selected start-goal pairs in the simulated environment. Table~I reports the average performance over 20 trials. GS-PRM+SLP reduced planned trajectory length, traveled path length, and navigation time by 5.16\%, 9.59\%, and 17.47\%, respectively.}

%We next compared GS-PRM+SLP with SplatNav~\cite{chen2025splat} in the same simulation environment. Both methods were given the same navigation task, and their planned and executed trajectories are shown in Fig.~\ref{fig:sim_navigation_compare}.

\begin{table}[t]
    \centering
    \renewcommand{\arraystretch}{1.15}
    \setlength{\tabcolsep}{8pt}
    \begin{tabular}{lcc}
        \hline
        \textbf{Metric} & \textbf{GS-PRM+SLP} & \textbf{SplatNav} \\
        \hline
        Planned trajectory length ($\mathrm{m}$)  & 37.474 & 39.512 \\
        Traveled trajectory length ($\mathrm{m}$) & 35.333 & 39.081 \\
        Travel time ($\mathrm{s}$)       & 166.733 & 202.033 \\
        \hline
    \end{tabular}
    \caption{\textbf{Navigation comparison between GS-PRM and SplatNav in simulation.}}
    \label{tab:round1_navigation_comparison}
    \vspace{-4mm}
\end{table}
% \vspace{-8pt}

% As summarized in Table~\ref{tab:round1_navigation_comparison}, both methods successfully reached the final goal. GS-PRM+SLP reduced the planned path length by $5.16\%$, the recorded travel distance by $9.59\%$, and the navigation time by $17.47\%$ compared with SplatNav.

\begin{figure}[t]
    \centering
    \includegraphics[width=0.7\linewidth]
    {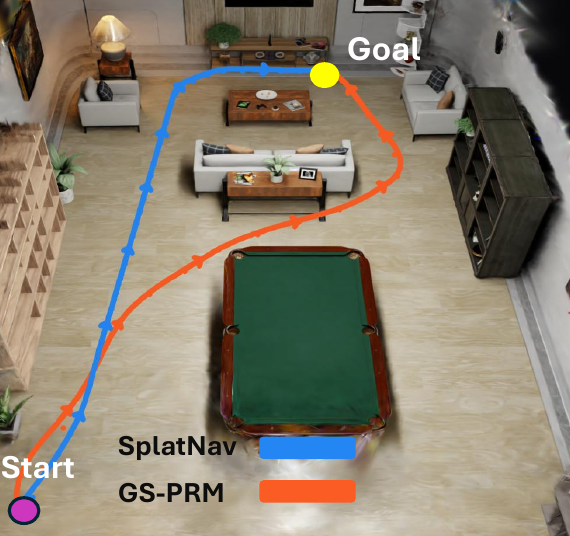}
    \caption{\textbf{Navigation comparison in the simulated environment.} The orange and blue lines show the planned trajectory using GS-PRM and SplatNav.}
    \label{fig:sim_navigation_compare}
    \vspace{-4mm}
\end{figure}

\subsection{Real-World Experiment }
We deployed GS-PRM and conducted online navigation experiments in two real-world environments. The experiments were designed to answer the following questions:

\begin{enumerate}
\item[\textbf{Q3}] Are images rendered from real-world 3DGS scenes useful for multi-query and long-horizon visual navigation?

\item[\textbf{Q4}]  Can GS-PRM serve as a global planner for existing goal-conditioned local navigation policies using images rendered from real-world 3DGS scenes?

\item[\textbf{Q5}]  How does the resulting navigation system compare with a state-of-the-art 3DGS-based navigation system?
\end{enumerate}

%\begin{figure}[htb!]
%\vspace*{-5pt}
%\centering
%\includegraphics[width=0.5\columnwidth]{figures/robot.pdf}
%\vspace*{-3pt}
%\caption{The robot and its driver PC used for the %experiments. %\blue{Note that the 2D LiDAR sensor is not used for navigation, but only for collision data collection and quantitative evaluation of baseline methods.} 
%}\vspace*{-8pt}
%\label{fig:former_robot}
%\end{figure}

%\noindent{\bf Roadmap construction:}
%For the T1-5 scene containing $630 K$ Gaussians, GS-PRM requires $32.09 \mathrm{s}$ for its construction. 
%additional offline computation on an NVIDIA RTX 5090 to construct a roadmap with 40 accepted vertices and 120 edges, achieving $98.09\%$ workspace coverage. The average computation time was \(86.1\,\mathrm{ms}\) per vertex trial and \(50.7\,\mathrm{ms}\) per edge trial.

\noindent {\bf Evaluation metrics:}
To quantify navigation performance, we evaluated each navigation task based on the following four measures:
\begin{enumerate}
\item Success Rate (SR)
\item SPL, success rates weighted by path length. %\footnote{For SPL evaluation, a reference path was computed using A* on a 2D occupancy grid map constructed via 2D-LiDAR SLAM.}
%\footnote{The global plan generated by GS-PRM was used as the reference path for computing SPL for all methods.}
\item Penalized Travel Time (PTT), defined as the total travel time reciprocal to the binary success indicator.
% \item Success-Weighted Time Efficiency (SWTE), defined as $\mathrm{SWTE}=S T_{\mathrm{ref}}/\max(T,T_{\mathrm{ref}})$, where $S$ is the binary success indicator, $T$ is the actual travel time, and $T_{\mathrm{ref}}=L_{\mathrm{ref}}/v_{\mathrm{ref}}$ is the reference travel time. Here, $L_{\mathrm{ref}}$ denotes the global path length, and $v_{\mathrm{ref}}$ denotes the reference robot speed\footnote{We set $v_{\mathrm{ref}}=0.2$ m/s for all experiments.}.
\item Global Path Segment Coverage (PC), which measures the ratio of the number of covered path segments to the total number of path segments over a multi-query navigation task. %, {\em i.e.,} $\mathrm{PC}=N_{\mathrm{cov}}/N_{\mathrm{tot}}$. %SPL and PTT can be less informative for unsuccessful episodes, particularly over long navigation horizons, since they do not reflect how far the robot progresses before failure. PC complements these metrics by quantifying the fraction of the global path successfully traversed. %sequences.
\end{enumerate}

\noindent {\bf Real-world environment setup:}
We arranged two sites for the navigation evaluations. The first environment (T1-5) is a cluttered laboratory studio characterized by rich background textures and relatively consistent illumination. This scene requires $630K$
Gaussians and takes $32s$ to construct GS-PRM.
The second environment (T2-5) is a corridor in a university building and is more challenging than T1-5 due to its relatively texture-poor surroundings, varying illumination conditions, and much longer navigation horizon. For each environment, given start and goal images, we solve the global path $\mathcal{T}_g \subset \mathcal{M}_{\mathrm{RC}}$ using GS-PRM, each consisting of five sub-paths. Each node of $\mathcal{T}_g$ also contains a sequence of rendered images $\{I_k^{\mathrm{ref}}\}$ to be used as subgoal images, as shown in Fig.~\ref{fig:topomap_envs}.

% T2-5 contains 427K Gaussians and requires $63.14s for roadmap construction because its \(327.84\,\mathrm{m}^2\) planning area is approximately \(4.60\times\) larger than the \(71.25\,\mathrm{m}^2\) area of T1-5.

\begin{figure*}[htb!]
\centering

% Row 1: Topomaps
\begin{subfigure}[c]{0.35\textwidth}
    \centering
    \includegraphics[width=\linewidth]
    {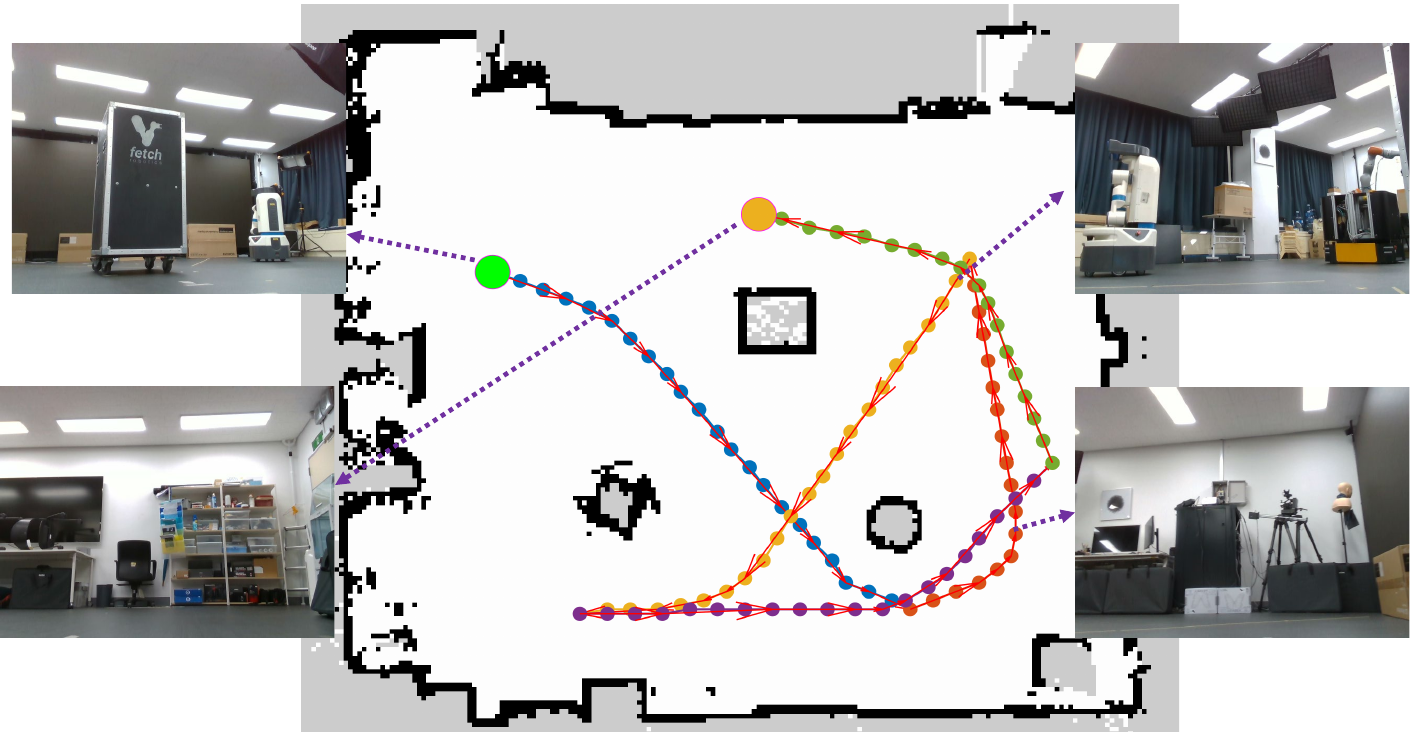}
    \caption{T1-5}
    \label{fig:TB1}
\end{subfigure}
\hfill
\begin{subfigure}[c]{0.64\textwidth}
    \centering
    \includegraphics[width=\linewidth]
    {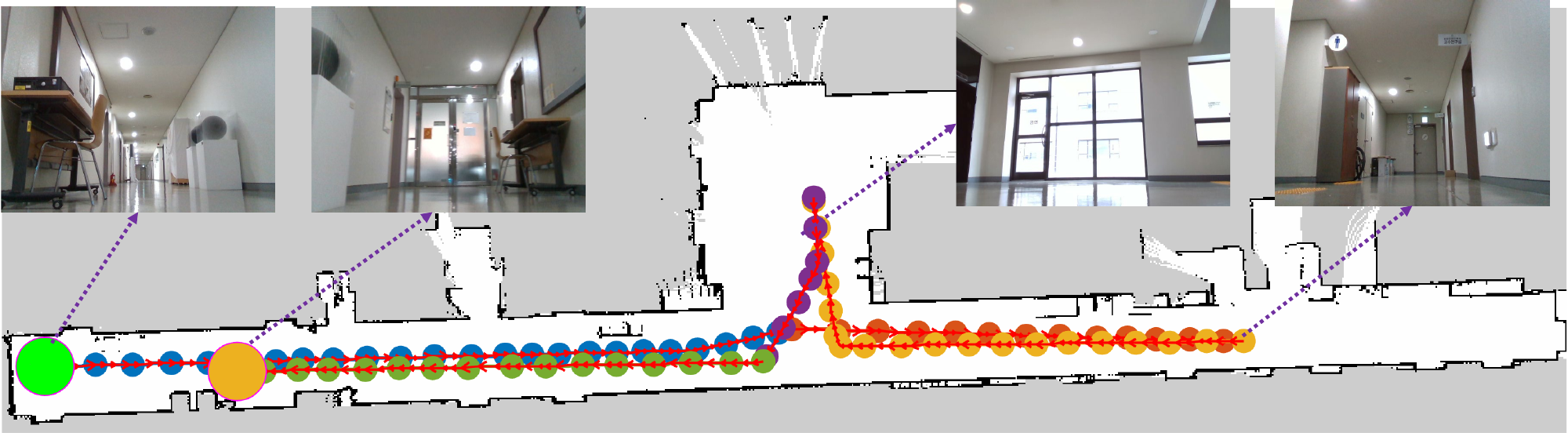}
    \caption{T2-5}
    \label{fig:TB2}
\end{subfigure}

%\vspace{-1mm}

% Row 2: Common legend
\includegraphics[width=0.65\textwidth]
{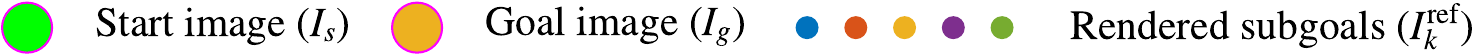}
    %\vspace{-2pt}
\caption{\textbf{Trajectories used in the experiments.}
The path lengths from the start to the final goal in (a) and (b) are 27.44~m and 72.99~m, respectively.}
\label{fig:topomap_envs}
\end{figure*}

%\vspace{-2pt}
% T1-5 results: 1 x 4, full two-column width
\begin{figure*}[!htb]
    \centering
    \captionsetup[subfigure]{font=footnotesize}
    \begin{subfigure}[t]{0.16\textwidth}
        \centering
        \includegraphics[width=\linewidth]
        {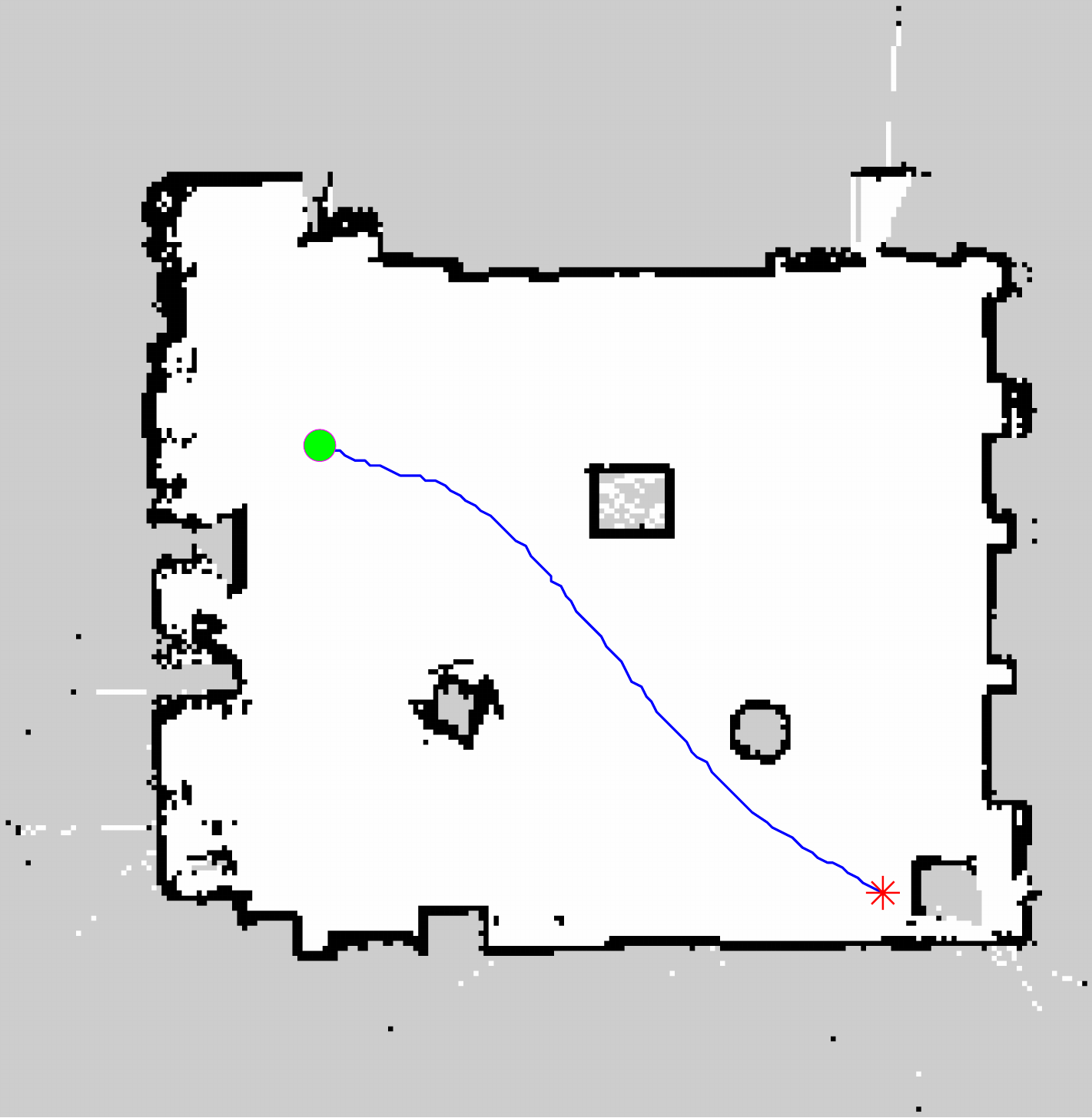}
        \caption{DevGRU}
        \label{fig:devgru_only-5-1}
    \end{subfigure}
    \hfill
    \begin{subfigure}[t]{0.16\textwidth}
        \centering
        \includegraphics[width=\linewidth]
        {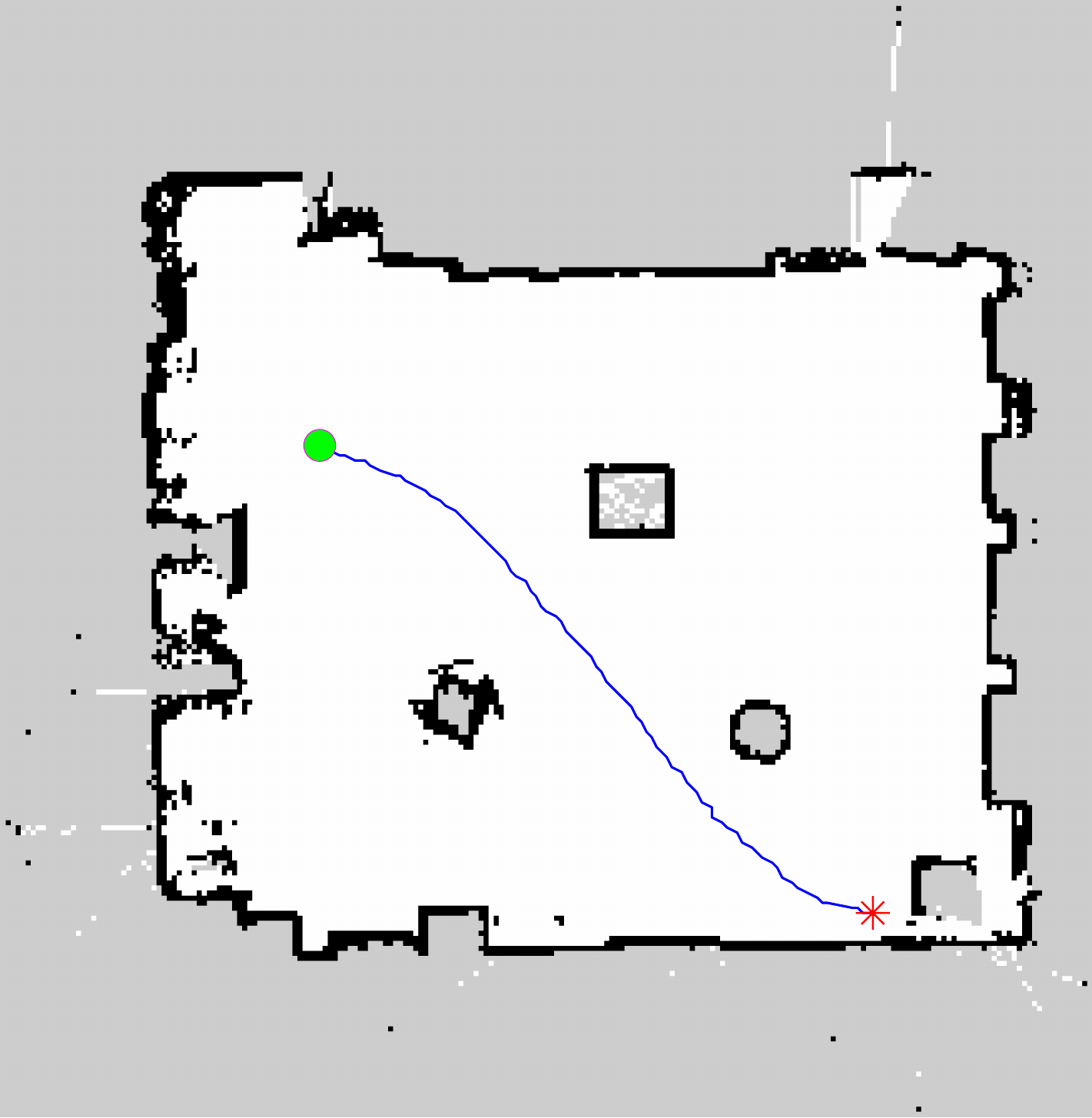}
        \caption{ViNT}
        \label{fig:vint_only-5-1}
    \end{subfigure}
    \hfill
    \begin{subfigure}[t]{0.16\textwidth}
        \centering
        \includegraphics[width=\linewidth]
        {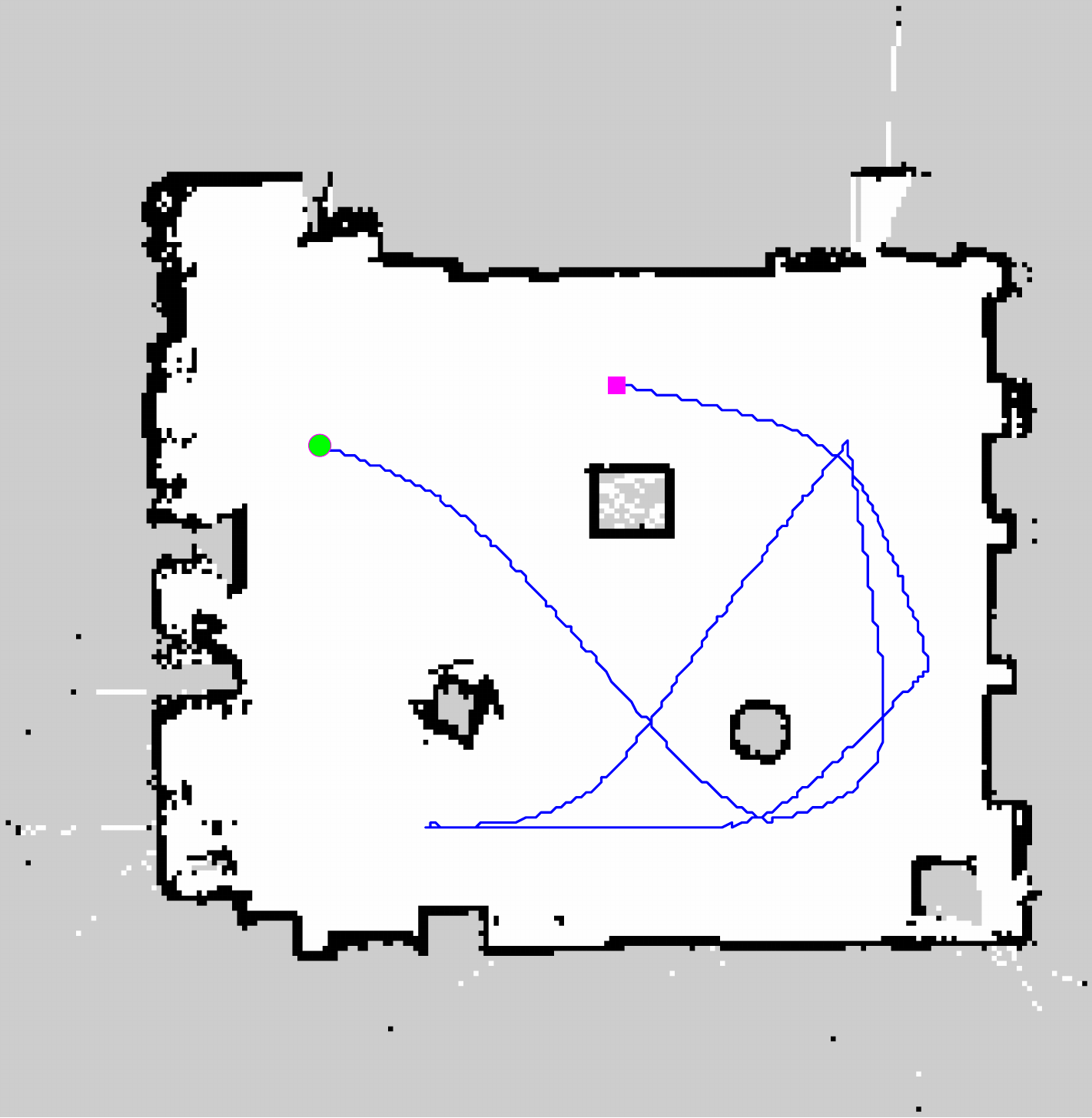}
        \caption{GS-PRM+SLP}
        \label{fig:method3-5-1}
    \end{subfigure}
    \hfill
    \begin{subfigure}[t]{0.16\textwidth}
        \centering
        \includegraphics[width=\linewidth]
        {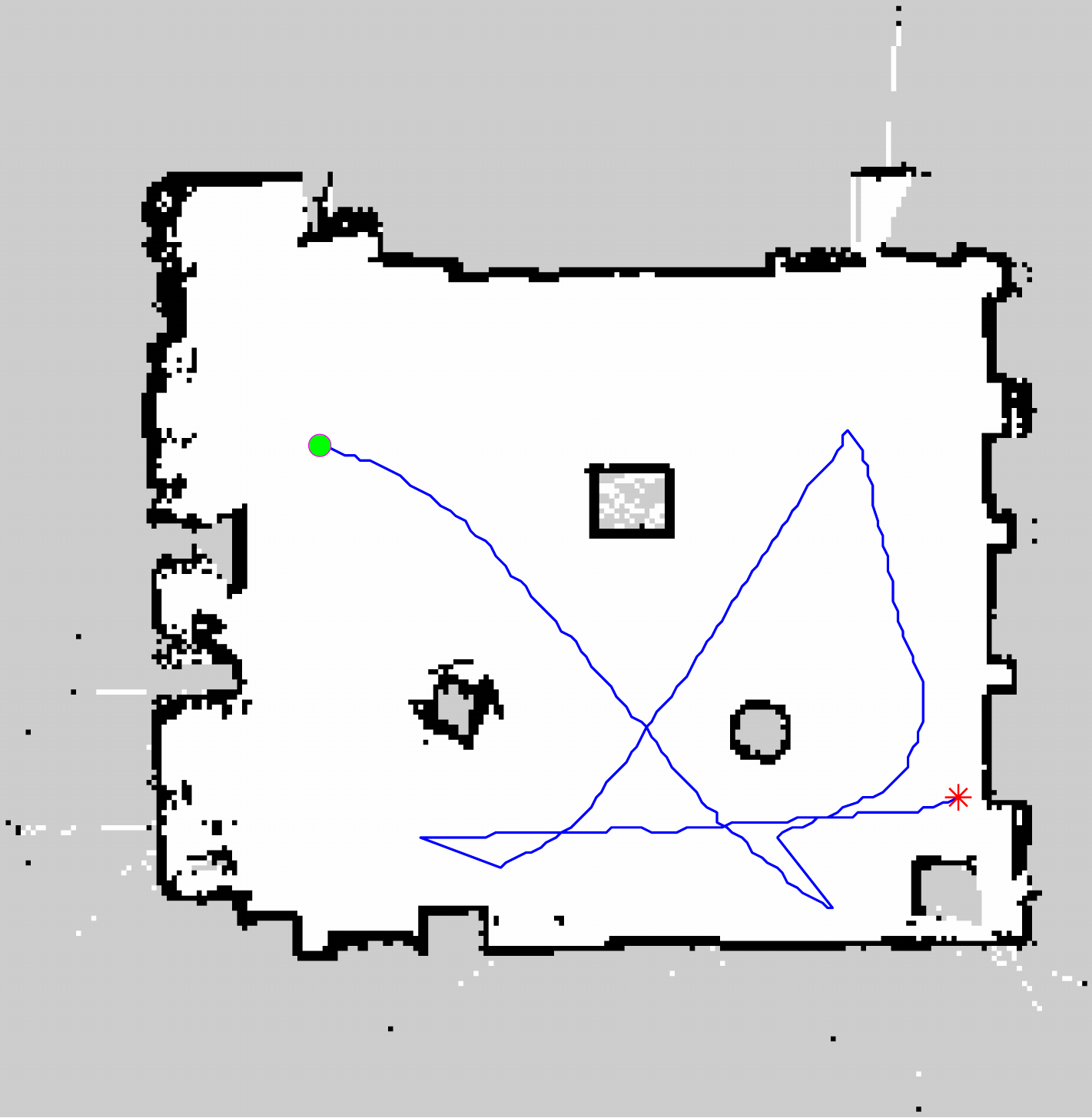}
        \caption{GS-PRM+ViNT}
        \label{fig:vint-5-1}
    \end{subfigure}
    \hfill
    \begin{subfigure}[t]{0.16\textwidth}
        \centering
        \includegraphics[width=\linewidth]
        {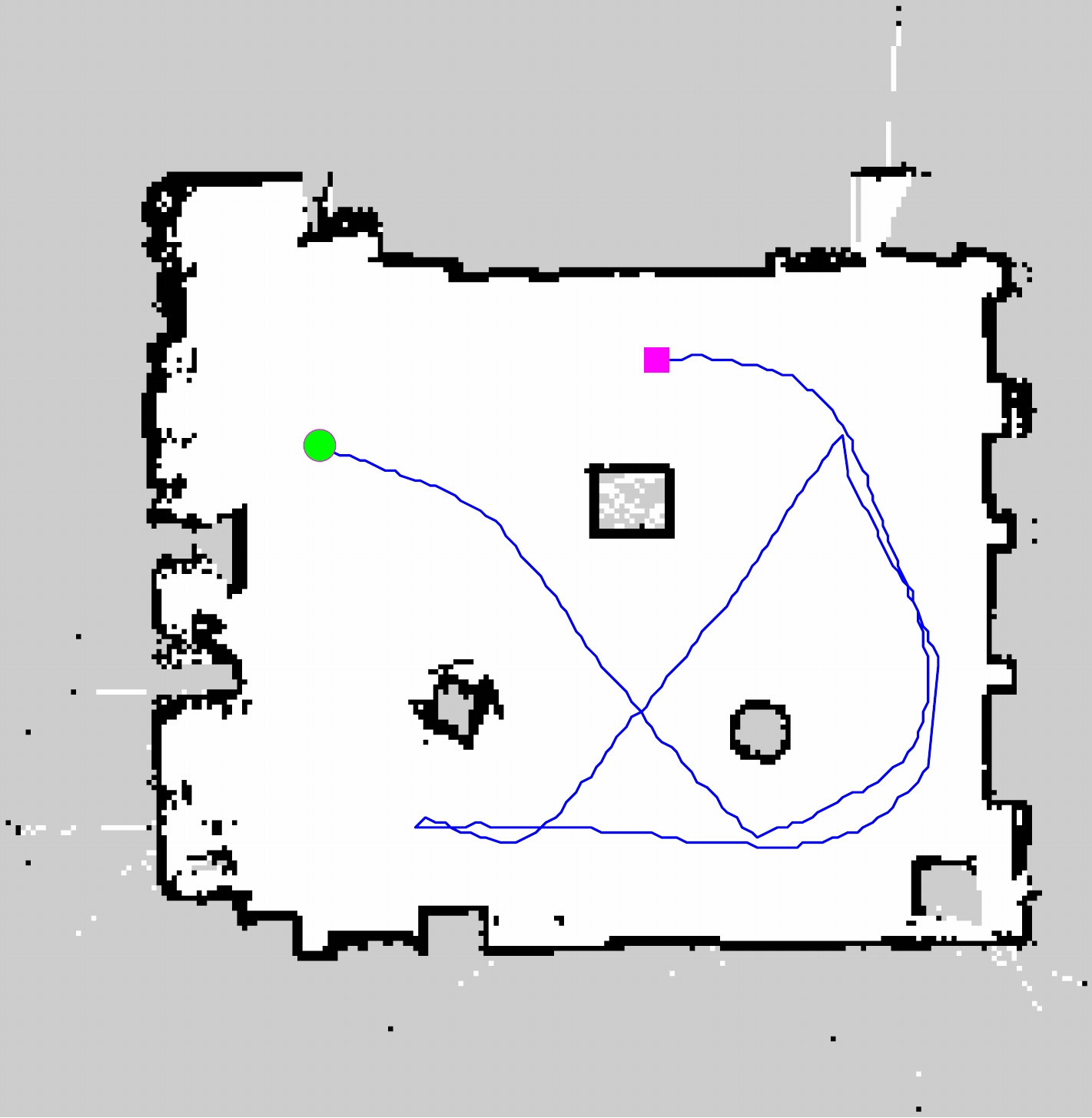}
        \caption{GS-PRM+DevGRU}
        \label{fig:devgru-5-1}
    \end{subfigure}
    \hfill
    \begin{subfigure}[t]{0.16\textwidth}
        \centering
        \includegraphics[width=\linewidth]
        {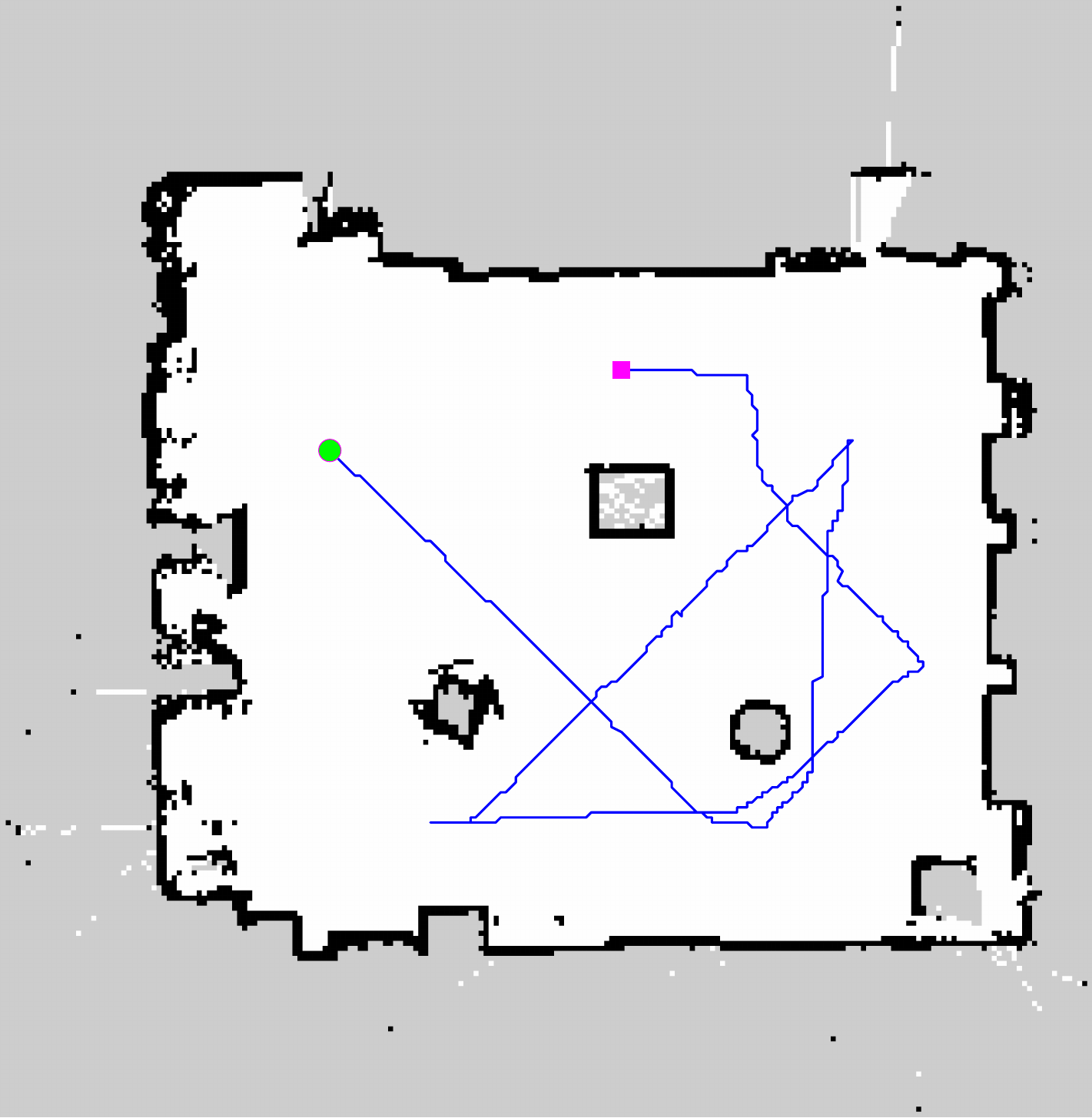}
        \caption{SplatNav}
        \label{fig:splatnav-5-1}
    \end{subfigure}

    \vspace{1mm}

    \includegraphics[width=0.8\textwidth]
    {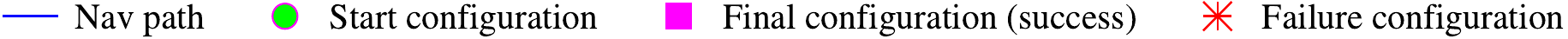}
    %\vspace{-1pt}
    \caption{\textbf{Navigation results in the T1-5 environment.}}
    \label{fig:T1-5_results}
\end{figure*}

%\vspace{-2pt}
% T2-5 results: 2 x 2, full two-column width
\begin{figure*}[!t]
    \centering

    % Row 1
    \begin{subfigure}[t]{0.32\textwidth}
        \centering
        \includegraphics[width=\linewidth]
        {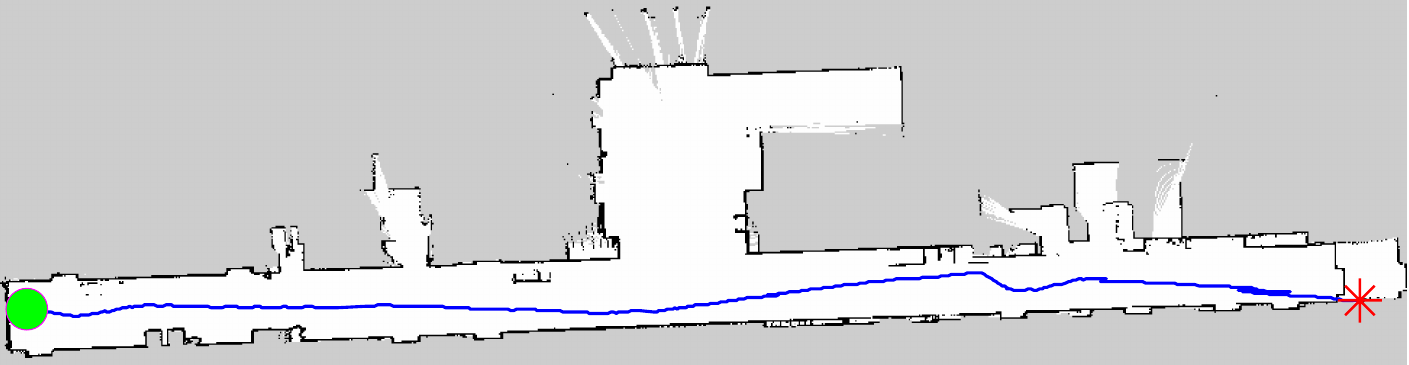}
        \caption{DevGRU}
        \label{fig:devgru_only1-5-2}
    \end{subfigure}
    \hfill
    \begin{subfigure}[t]{0.32\textwidth}
        \centering
        \includegraphics[width=\linewidth]
        {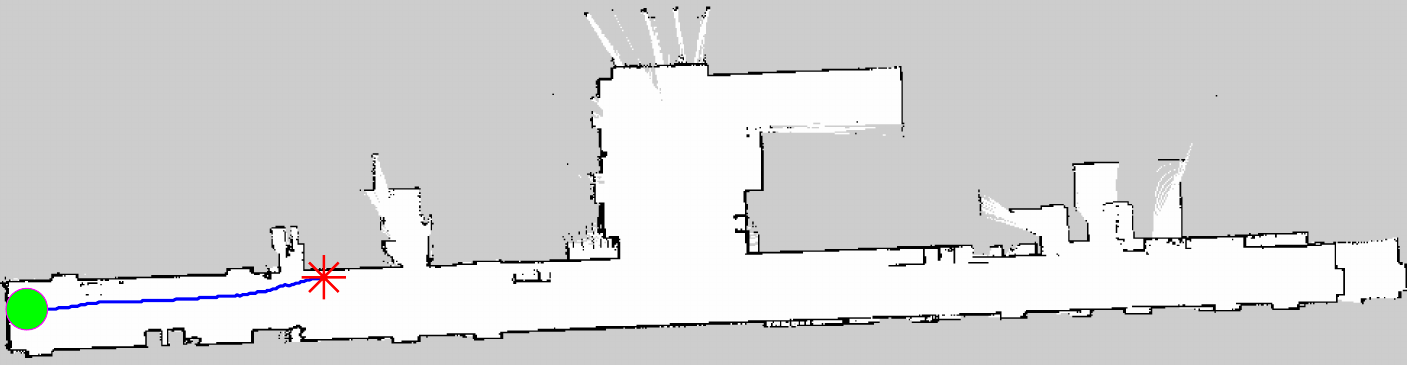}
        \caption{ViNT}
        \label{fig:vint_only-5-2}
    \end{subfigure}
    \hfill
    \begin{subfigure}[t]{0.32\textwidth}
        \centering
        \includegraphics[width=\linewidth]
        {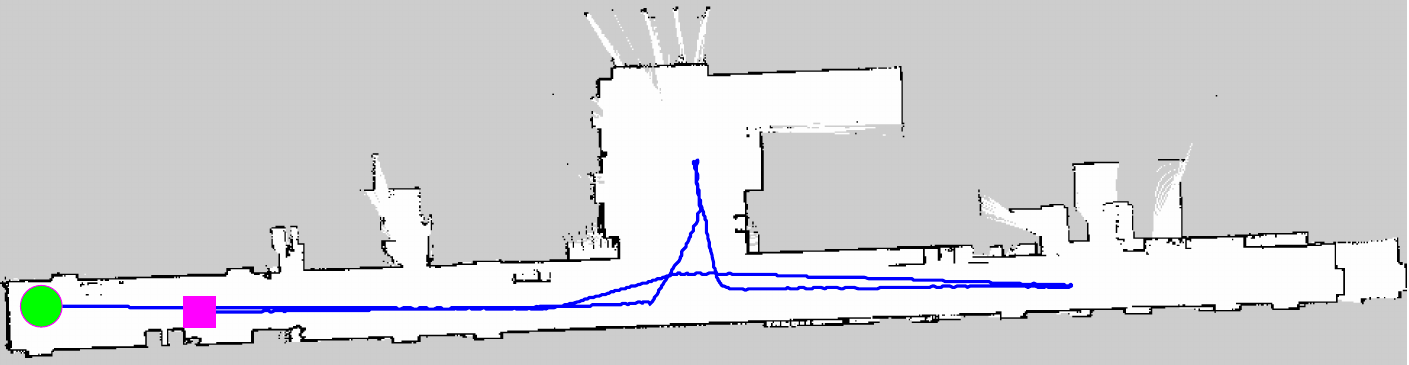}
        \caption{GS-PRM+SLP}
        \label{fig:SL-5-2}
    \end{subfigure}

    \par\vspace{1mm}

    % Row 2
    \begin{subfigure}[t]{0.32\textwidth}
        \centering
        \includegraphics[width=\linewidth]
        {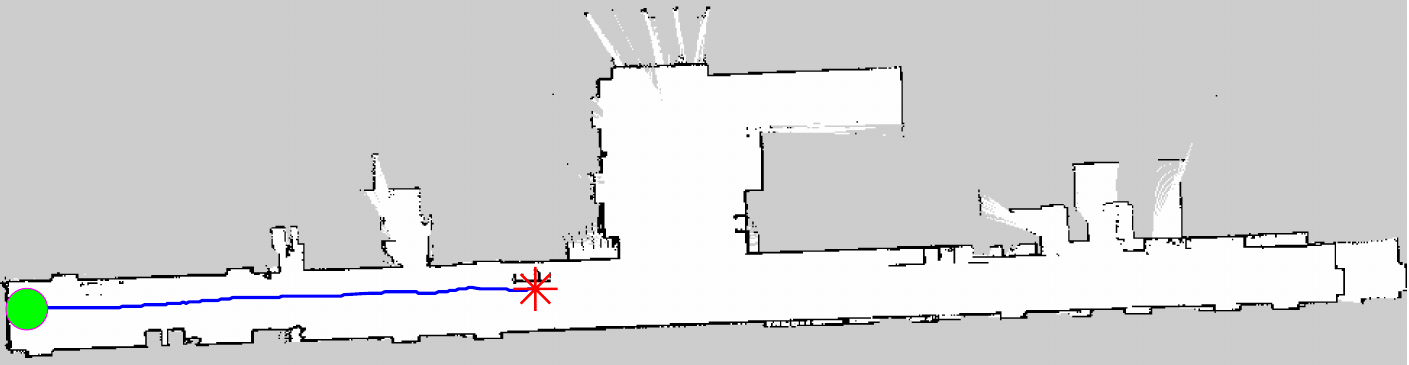}
        \caption{GS-PRM+ViNT}
        \label{fig:vint-5-2}
    \end{subfigure}
    \hfill
    \begin{subfigure}[t]{0.32\textwidth}
        \centering
        \includegraphics[width=\linewidth]
        {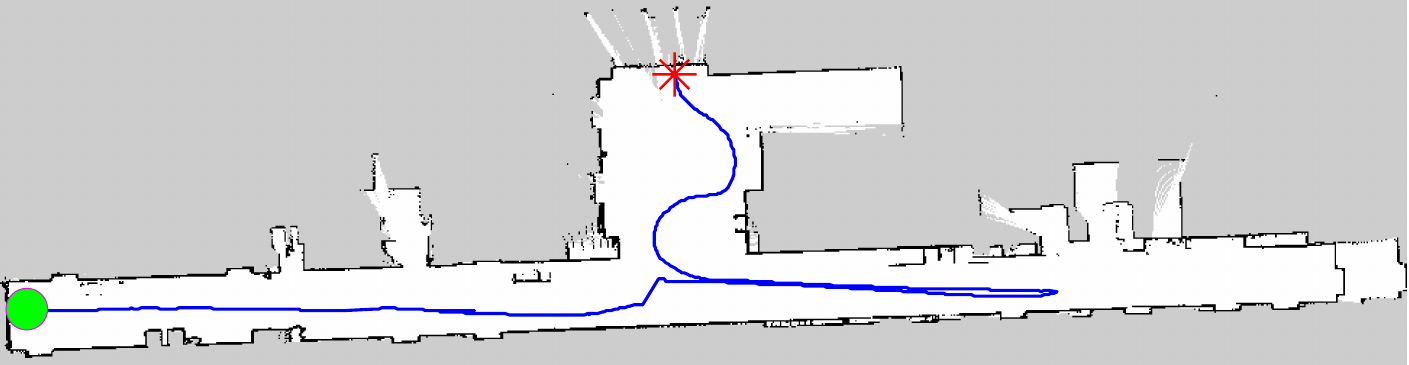}
        \caption{GS-PRM+DevGRU}
        \label{fig:devgru-5-2}
    \end{subfigure}
    \hfill
    \begin{subfigure}[t]{0.32\textwidth}
        \centering
        \includegraphics[width=\linewidth]
        {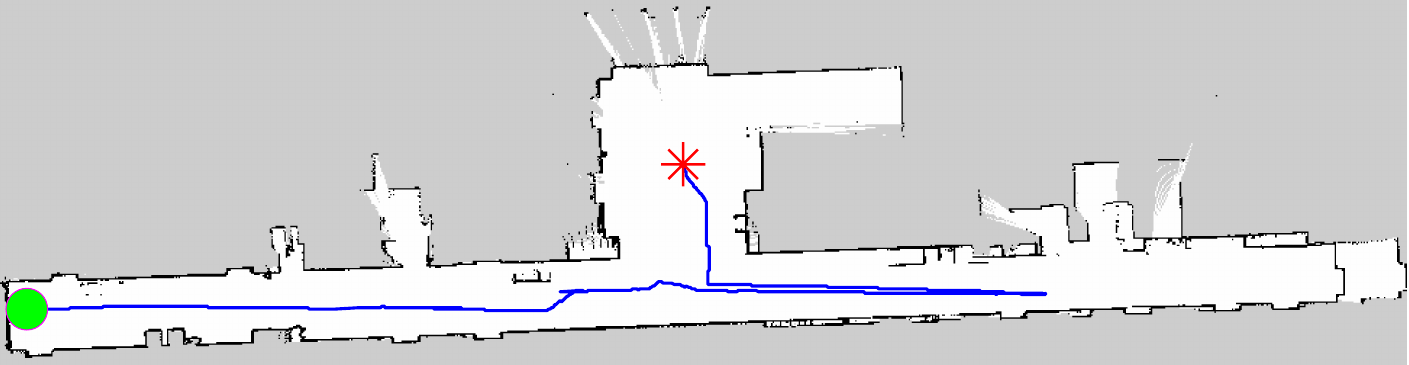}
        \caption{SplatNav}
        \label{fig:splatnav-5-2}
    \end{subfigure}

    \par\vspace{2mm}

    \includegraphics[width=0.8\textwidth]
    {figures/exp_res/res_legend_1row.pdf}

    \caption{\textbf{Navigation results in the T2-5 environment.}}
    \label{fig:T2-5_results}
\end{figure*}

\begin{table*}[!t]
\centering
\scriptsize
\setlength{\tabcolsep}{3pt}
\renewcommand{\arraystretch}{1.15}

\resizebox{0.90\textwidth}{!}{%
\begin{tabular}{l|cccc|cccc|cccc}
\toprule[1.5pt]

% \multirow{Method}
\multirow{2}{*}{\textbf{Method}}
& \multicolumn{4}{c|}{T1-5}
& \multicolumn{4}{c|}{T2-5}
& \multicolumn{4}{c}{Average} \\

\cmidrule(lr){2-5}
\cmidrule(lr){6-9}
\cmidrule(lr){10-13}

& SR $\uparrow$
& PC $\uparrow$
& SPL $\uparrow$
& PTT (s) $\downarrow$
& SR $\uparrow$
& PC $\uparrow$
& SPL $\uparrow$
& PTT (s) $\downarrow$
& SR $\uparrow$
& PC $\uparrow$
& SPL $\uparrow$
& PTT (s) $\downarrow$ \\

\midrule

DevGRU (Standalone)
& 0.00 & 0.20 & 0.00 & $-$
& 0.00 & 0.40 & 0.00 & $-$
& 0.00 & 0.30 & 0.00 & $-$ \\

ViNT (Standalone)
& 0.00 & 0.20 & 0.00 & $-$
& 0.00 & 0.00 & 0.00 & $-$
& 0.00 & 0.10 & 0.00 & $-$ \\

\midrule

GS-PRM+SLP
& \bf{1.00} & \bf{1.00} & 0.98 & \bf{136.02}
& \bf{1.00} & \bf{1.00} & \bf{0.99} & \bf{260.40}
& \bf{1.00} & \bf{1.00} & \bf{0.99} & \bf{198.21} \\

GS-PRM+ViNT
& 0.00 & 0.60 & 0.00 & $-$
& 0.00 & 0.00 & 0.00 & $-$
& 0.00 & 0.30 & 0.00 & $-$ \\

GS-PRM+DevGRU
& \bf{1.00} & \bf{1.00} & 0.93 & 228.78
& 0.00 & 0.40 & 0.00 & $-$
& 0.50 & \bf{0.70} & 0.46 & $-$ \\

SplatNav
& \bf{1.00} & \bf{1.00} & \bf{1.00} & 317.97
& 0.00 & 0.40 & 0.00 & $-$
& 0.50 & \bf{0.70} & 0.50 & $-$ \\

\bottomrule[1.5pt]
\end{tabular}%
}

\caption{\textbf{Navigation performance in the T1-5 and T2-5 environments.} 
The top two standalone local planners were tested without exploiting GS-PRM 
for visual localization and pose correction, whereas the bottom three did. 
The reported values are averaged over two runs for each method in each environment.}
\label{tab:nav_performance}
\vspace{-4mm}
\end{table*}

%\vspace{-8pt}

%-------------------------------------------------------------------------
% \subsection{Validation of 3DGS-Based Collision Checking} \label{sec:collision_validation}

% We evaluated the proposed 3DGS-based collision checker in a customized Isaac Sim environment with known collision geometry. The 3DGS scene was reconstructed from simulator-rendered images and used to classify uniformly sampled robot positions as RC-free or rejected. For ground-truth validation, the robot was placed at each evaluated position in the original Isaac Sim scene, and collisions were determined using the simulator geometry. Floor contacts below the specified collision-height threshold were ignored.

% A total of 654 samples were evaluated. All 131 samples accepted by the 3DGS checker were collision-free in Isaac Sim. Of the 523 rejected samples, 334 were confirmed collisions and 189 were collision-free. The latter cases mainly reflect the conservative RC-free criterion, which also considers robot footprint and safety clearance rather than physical contact alone.

\subsubsection{Navigation Performance with Different Local Planning Policies}
To answer Q3--Q4, we conducted navigation experiments in T1-5 and T2-5. 
Using $\mathcal{T}_g$, we employed three different local navigation policies: a straight-line planner (SLP), ViNT~\cite{shah2023vint}, and DevGRU~\cite{han2026devgru}. Note that SLP provides a straightforward means of evaluating the utility of rendered images for visual navigation, thereby addressing Q3.
%For each online navigation task, the robot starts at the designated start vertex, follows the global plan by sequentially navigating and exploiting the rendered subgoal images, and completes the task upon reaching the final goal vertex.

% Fig.~\ref{fig:T1-5_results} shows the navigation results in the T1-5 environment. All methods successfully completed the navigation task except ViNT. One possible reason for ViNT's failure is the domain gap between the rendered images and the real-world images used to train the policy, as ViNT relies solely on RGB observations for visual navigation. DevGRU may also be affected by a similar domain gap in the rendered depth images. Nevertheless, DevGRU successfully completed the task, potentially because it additionally utilizes the relative subgoal (SG) position as an input. In contrast, T1-5 was relatively straightforward for SLP because it is designed to directly exploit the rendered observations for frequent pose estimation and correction during navigation without relying a learned navigation policy. 

{Fig.~\ref{fig:T1-5_results} shows the navigation results in the T1-5 environment. Employing the goal-conditioned policies, DevGRU and ViNT, as standalone policies resulted in unreliable navigation. In contrast, the three GS-PRM-guided policies and SplatNav, which actively exploit 3DGS rendering for pose correction or planning, generally achieved more reliable navigation. Among them, GS-PRM+ViNT still struggled with the navigation task. One possible reason is the domain gap between the rendered images and the real-world images used to train ViNT, as ViNT relies solely on RGB observations for visual navigation. In contrast, T1-5 was relatively straightforward for SLP because it directly exploits rendered observations for frequent pose estimation and correction during navigation, without relying on a learned navigation policy.}

Fig.~\ref{fig:T2-5_results} shows the navigation results in the T2-5 environment. Unlike in T1-5, all methods except the SLP failed to complete the full navigation task. Successful navigation with the SLP demonstrates that the rendered observations and global plan generated by GS-PRM can support online navigation, thereby providing a positive answer to Q3. The failures of the two goal-conditioned local policies, ViNT and DevGRU, are likely associated with the domain gap between their training data and the rendered observations. 
%Nevertheless, their failure tendency was consistent with that observed in T1-5: DevGRU progressed farther along the global plan than ViNT, presumably because it additionally exploits relative subgoal position information. 
These results suggest that, with respect to Q4, GS-PRM can serve as a global planner for goal-conditioned local navigation policies, while the resulting navigation performance can depend on the quality of the reconstructed 3DGS scene and the local policy's robustness to rendered observations. This may also explain why the simulation-only experiments reported in~\cite{honda2025gsplatvnm} achieved more successful navigation results. %, where the discrepancy between rendered observations and the observations encountered during navigation is expected to be smaller.

\subsubsection{Comparison against SplatNav}
To answer Q5, we conducted additional experiments with SplatNav~\cite{chen2025splat}, a state-of-the-art 3DGS-based navigation system, in both T1-5 and T2-5. SplatNav employs a simple trajectory controller similar to our SLP, while its global plan is generated by a separate planning procedure called Splat-Plan. As shown in Figs.~\ref{fig:splatnav-5-1} and~\ref{fig:splatnav-5-2} and Table~\ref{tab:nav_performance}, SplatNav performed comparably to GS-PRM+DevGRU and worse than GS-PRM+SLP. In T1-5, its travel time was approximately 39\% and 134\% longer than those of GS-PRM+DevGRU and GS-PRM+SLP, respectively. This is mainly attributed to its polytope-based acceleration mechanism, which is less effective in environments with narrow passages and cluttered obstacles.

% \begin{figure}[t] 
% \centering 
% \includegraphics[width=\linewidth] 
%     {figures/splatnav_GS-PRM_compare.pdf}
%     \caption{\label{fig:pipeline}
% {Comparison of navigation using (a) RV-PRM and (b) SplatNav. Dotted and solid lines denote the planned and executed trajectories, respectively, with colors identifying individual trajectory segments.} }
% \label{fig:pipeline}
% \end{figure}

% \begin{table}[t]
%     \centering
%     \caption{Comparison between GS-PRM+SLP and SplatNav in simulation environment.}
%     \label{tab:round1_navigation_comparison}
%     \begin{tabular}{lcc}
%         \hline
%         \textbf{Metric} & \textbf{GS-PRM+SLP} & \textbf{SplatNav} \\
%         \hline
%         Planned distance ($\mathrm{m}$)  & 37.474 & 39.512 \\
%         Recorded distance ($\mathrm{m}$) & 35.333 & 39.081 \\
%         Navigation time ($\mathrm{s}$)   & 166.733 & 202.033 \\
%         Average speed ($\mathrm{m\,s^{-1}}$) & 0.212 & 0.193 \\
%         Final goal reached               & Yes & Yes \\
%         \hline
%     \end{tabular}
% \end{table}

% We compared GS-PRM+SLP with SplatNav in the simulation environment, as summarized in Table~\ref{tab:round1_navigation_comparison}. Both methods successfully reached the final goal. GS-PRM+SLP produced a shorter planned and executed trajectory and completed the navigation faster than SplatNav, reducing the planned path length by $5.16\%$, the recorded travel distance by $9.59\%$, and the navigation time by $17.47\%$.

%\section{Limitations}

\section{Conclusion}

We presented \method{}, a coverage-induced visual roadmap constructed through
3DGS rendering. RC-free distinguishes render certification from physical
collision freedom and states the assumptions linking them. Under conservative
rendering and robust local-connectability conditions, uniform workspace
sampling is probabilistically complete for robust RC-free queries; once
certified coverage and connectivity hold, query completeness is deterministic.
The roadmap also supports image-defined queries and rendered visual subgoals.
Validating the render oracle, therefore, remains as important as evaluating
connectivity and navigation performance. There are a few limitations to GS-PRM that we would like to address in the future. Sampling in unknown or poorly reconstructed regions remains challenging because the RC-free criterion relies on sufficient rendering support from the 3DGS representation. The completeness guarantee also depends on the reconstruction quality.
Our current formulation assumes a static environment and does not explicitly account for dynamic obstacles or scene changes.

\bibliographystyle{IEEEtran}
\bibliography{references}

\end{document}